\documentclass[lettersize,journal]{IEEEtran}
\usepackage{amsmath,amsfonts}
\usepackage{algorithmic}
\usepackage{array}
\usepackage[caption=false,font=normalsize,labelfont=sf,textfont=sf]{subfig}
\usepackage{textcomp}
\usepackage{stfloats}
\usepackage{url}
\usepackage{verbatim}
\usepackage{graphicx}

\graphicspath{{figures/}}
\usepackage{xcolor}
\usepackage[dvipsnames,table]{xcolor}
\usepackage{colortbl}
\usepackage{booktabs}
\usepackage{multirow}
\usepackage{makecell}
\usepackage{threeparttable}
\usepackage{tabulary}
\usepackage{rotating}
\usepackage{tabularx}
\usepackage[hidelinks]{hyperref}

\def\BibTeX{{\rm B\kern-.05em{\sc i\kern-.025em b}\kern-.08em
    T\kern-.1667em\lower.7ex\hbox{E}\kern-.125emX}}
\usepackage{balance}
\begin{document}
%\title{Learning Locally Aligned Visual Representations from Radiology Reports for Chest X-ray Vision--Language Models}
\title{Learning Entity-Informed Token Weights for Local Vision--Language Pretraining in Chest X-rays}
%\author{IEEE Publication Technology Department
%\thanks{Manuscript created October, 2020; This work was developed by the IEEE Publication Technology Department. This work is distributed under the \LaTeX \ Project Public License (LPPL) ( http://www.latex-project.org/ ) version 1.3. A copy of the LPPL, version 1.3, is included in the base \LaTeX \ documentation of all distributions of \LaTeX \ released 2003/12/01 or later. The opinions expressed here are entirely that of the author. No warranty is expressed or implied. User assumes all risk.}}
%
%\markboth{Journal of \LaTeX\ Class Files,~Vol.~18, No.~9, September~2020}%
%{How to Use the IEEEtran \LaTeX \ Templates}
\author{Kang Liu, Zhuoqi Ma \IEEEmembership{Member, IEEE}, Kun Xie, Yunan Li \IEEEmembership{Member, IEEE}, Ruyi Liu, Ronghan Li, Guang Jia, and Qiguang Miao \IEEEmembership{Senior Member, IEEE} 
	\thanks{All authors are with the School of Computer Science and Technology, Xidian University, Xi'an, Shaanxi 710071, China; Xi'an Key Laboratory of Big Data and Intelligent Vision, Xi'an, Shaanxi 710071, China; and Key Laboratory of Collaborative Intelligence Systems, Ministry of Education, Xidian University, Xi'an 710071, China (Corresponding author: Qiguang Miao, e-mail: qgmiao@xidian.edu.cn).}
	\thanks{Kang Liu and Zhuoqi Ma contributed equally to this work. }
	\thanks{This manuscript is currently under peer review. Copyright may be transferred without notice, after which this version may no longer be accessible.}
}

\maketitle

\begin{abstract}
Medical vision–language models (VLMs) predominantly rely on global image--report alignment, which captures holistic semantic correspondence but provides limited supervision for localized visual representations. Methods that explicitly learn local representations often require costly spatial annotations, including bounding boxes, segmentation masks, or radiologist gaze. We present \textbf{CheXLoc}, a report-supervised framework that learns locally aligned visual representations from paired chest radiographs and reports without explicit spatial annotations. CheXLoc derives entity-informed token weights by learning importance weights for clinical entity types and mapping them to the corresponding report tokens. The resulting token weights modulate the contributions of individual tokens to local patch--token alignment, complementing global image--report alignment. Pretrained on 240,197 radiographs from 59,693 patients, CheXLoc achieves the highest 1/RadCliQ-V1 scores among the evaluated methods on MIMIC-CXR, CheXpert Plus, and ReXGradient. Relative to MAIRA-2-7B, its scores are higher by 53.2\%, 8.2\%, and 55.7\%, respectively. With 0.282 billion parameters, CheXLoc requires 17.4\% as much peak GPU memory and 22.7\% as much inference time as MAIRA-2-7B. Its visual representations also transfer to disease classification, achieving AUROCs of 0.966 and 0.937 on Shenzhen and TBX11K, respectively, with 1\% of the downstream training data. Image-removal and lesion-level perturbation analyses further show that generated reports are sensitive to visual input and disease-specific regions. These results support the use of radiology reports as scalable weak supervision for learning locally aligned visual representations without spatial annotations. Code will be available at https://github.com/mk-runner/CheXLoc.
\end{abstract}


\section{Introduction}
\IEEEPARstart{C}{hest} X-ray (CXR) imaging is one of the most widely used diagnostic modalities in clinical practice, providing critical information for detecting and monitoring thoracic abnormalities \cite{2025-eva-x,2025-nbe-pixel}. Recent advances in vision--language models (VLMs) have enabled joint learning from large-scale collections of paired radiographs and reports, facilitating a range of medical imaging tasks, including disease classification \cite{2026-nbe-afloc}, lesion segmentation \cite{2026-cogaze}, and report generation \cite{2025-nc-llava-rad,Liu-2025-CVPR-mlrg}. However, existing medical VLMs \cite{zhang2025biomedclipmultimodalbiomedicalfoundation,PriorRG} predominantly learn global image representations that summarize the radiograph as a whole, potentially overlooking fine-grained visual semantics associated with individual anatomical structures and pathological findings.

\begin{figure*}
	\centering
	\includegraphics[width=1\linewidth]{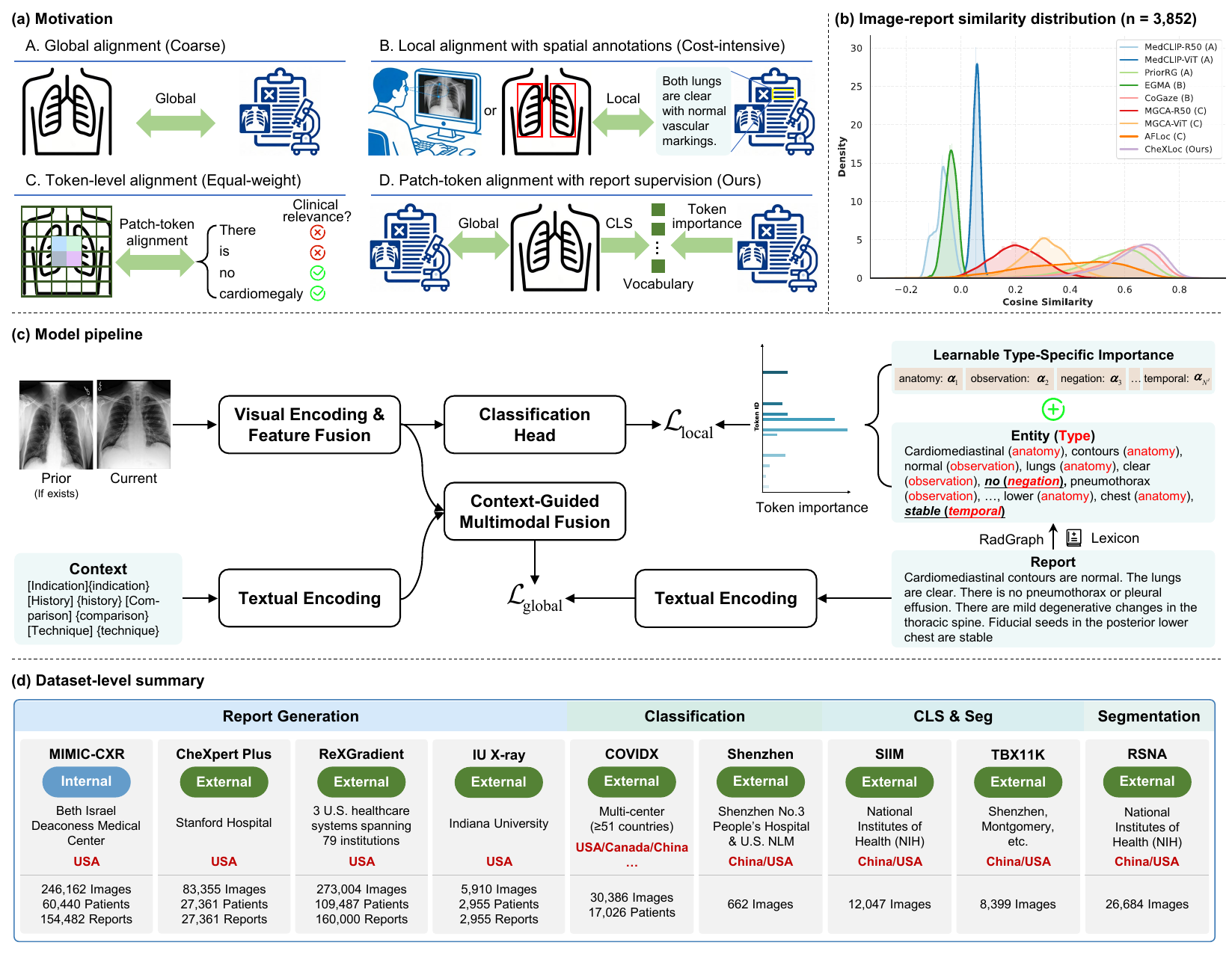}
	\caption{
		Overview of CheXLoc.
		\textbf{a,} Comparison of existing approaches and CheXLoc. Unlike existing methods that rely on uniform token weighting or spatial supervision, CheXLoc derives entity-informed token weights from report-level clinical entities to modulate local patch--token alignment without requiring spatial annotations.
		\textbf{b,} Image--report cosine similarity on the full MIMIC-CXR test set ($n = 3{,}852$). CheXLoc achieves the highest cosine similarity among the evaluated methods.
		\textbf{c,} Architecture of CheXLoc. Bold, italic, and underlined \textit{\textbf{\underline{entities}}} indicate entities omitted by RadGraph and subsequently recovered using an expert-curated lexicon.
		\textbf{d,} Dataset statistics for the study cohorts. Reported numbers represent the total size of each dataset. MIMIC-CXR comprises 246,162 images, of which 240,197 images from the official training split were used for pretraining.
	}
	\label{fig:overview}
\end{figure*}

An alternative is to learn fine-grained cross-modal correspondences directly from paired radiographs and reports, without explicit spatial annotations. MGCA \cite{wang-mgca} uses multi-granularity contrastive learning to align visual and textual representations at different semantic levels, while AFLoc~\cite{2026-nbe-afloc} explores hierarchical alignment between visual and textual features. These studies show that paired radiographs and reports can provide implicit supervision for local cross-modal alignment. However, their alignment objectives do not explicitly model differences in the clinical relevance of textual tokens to local visual evidence. As a result, clinically informative entities (e.g., \textit{opacity} and \textit{effusion}) may receive the same emphasis as function words and syntactic fillers during patch--token alignment (Fig.~\ref{fig:overview}a), potentially introducing noisy supervision.

Recent work has also explored token weighting based on corpus statistics. For example, SuperCLIP~\cite{zhao2025-superclip} estimates token importance using inverse document frequency. Token frequency, however, does not necessarily reflect clinical relevance to visual evidence: frequent anatomical terms may provide essential context, whereas less frequent findings and modifiers expressing negation or temporality may be equally important. Frequency-based weighting alone may therefore fail to capture the varying contributions of clinical tokens to local alignment.

% ==============motivation & overview ========================

These observations motivate our hypothesis that clinical entities in radiology reports can provide weak supervision for local visual representation learning by modeling the relative importance of textual tokens. In particular, the semantic types of clinical entities provide a basis for differentiating the contributions of report tokens to local alignment, beyond weighting tokens according to their corpus-level frequency. We therefore introduce \textbf{CheXLoc}, a report-supervised VLM that incorporates entity-informed token weighting into local patch--token alignment. Specifically, Specifically, CheXLoc learns type-specific importance weights for clinical entities and maps them to the corresponding tokens in each report. The resulting token weights are incorporated into the local alignment objective, together with an adaptive reweighting factor that accounts for visual prediction difficulty. In parallel, global image--report alignment is retained to capture holistic cross-modal correspondence. This formulation enables local visual--textual representation learning from paired radiographs and reports without requiring explicit spatial annotations.

We evaluate CheXLoc on three aspects of report-supervised representation learning. First, we assess CXR report generation on internal and external cohorts, including zero-shot transfer to an independent institution. Second, we evaluate transfer to disease classification and lesion segmentation, including limited-label settings for classification. Third, we examine sensitivity to visual evidence through image-removal and lesion-level perturbation analyses. Our contributions are as follows:
\begin{itemize}
	\item We introduce a report-supervised framework for learning locally aligned visual representations from naturally paired chest radiographs and radiology reports, without requiring explicit spatial annotations.

	\item We develop an entity-informed token weighting mechanism that learns type-specific importance weights from report-derived clinical entities and maps them to tokenizer tokens for weighted local patch--token alignment.
	
	\item We introduce a perturbation-based evaluation protocol for radiology report generation, assessing image-level visual dependence and proposing Localization-Aware Generation Evaluation (LAGE) to quantify sensitivity to lesion-specific evidence.
	
	\item We conduct extensive experiments on report generation, disease classification, and lesion segmentation, demonstrating the effectiveness of CheXLoc across cross-institutional and zero-shot settings.
\end{itemize}

\section{Related Work}

\subsection{Medical Vision--Language Models}
Medical vision--language models (VLMs) learn joint representations from paired medical images and reports. Some methods introduce additional supervision to improve visual representation learning. CoCa-CXR \cite{chen2025coca-cxr} uses large language models to extract temporal structures and models disease progression with scene graphs \cite{chest-imagenome-1.0.0}. DCXFM \cite{2025-tmi-foundation-model-bbox} incorporates bounding boxes, pathology masks, and disease labels. EGMA \cite{ma2024-nips-eye-egma} and CoGaze \cite{2026-cogaze} use radiologists' eye-tracking signals to guide attention toward diagnostically relevant regions. Although these signals provide spatial constraints, collecting them increases annotation costs and may limit scalability.

Other methods pursue fine-grained representation learning through multi-level or local alignment. MGCA~\cite{wang-mgca} uses multi-granularity contrastive learning, whereas MRM~\cite{zhou2023-iclr-MRM} and MaCo~\cite{huang2024-NC-MaCo} combine masked modeling with contrastive objectives. CheXWorld~\cite{2025-cvpr-chexworld} adopts the JEPA framework~\cite{2023-cvpr-jepa} to model local anatomical structures, global layouts, and domain variations. AFLoc~\cite{2026-nbe-afloc} aligns visual features at shallow, deep, and global levels with textual representations at word, sentence, and global levels. Our previous work, PriorRG~\cite{PriorRG}, explores prior-guided contrastive pretraining. SuperCLIP~\cite{zhao2025-superclip} estimates token importance using inverse document frequency. Unlike these methods, CheXLoc uses clinical entities to estimate token importance and modulate local patch--token alignment, without requiring additional spatial annotations.

\subsection{Chest X-ray Report Generation}
Chest X-ray report generation aims to produce clinically informative reports from radiographs. Knowledge-guided methods \cite{huang-kiut,shen2024automatic_aaai,AM-MRG-2026-tmi} incorporate structured medical knowledge or disease-related memory. Temporal methods \cite{PriorRG,Liu-2025-CVPR-mlrg,26-cvpr-biotprompt} use longitudinal images to model disease progression and patient-specific changes. Recent LLM-based approaches include PromptMRG \cite{promtmrg-aaai-2024}, R2-LLM \cite{aaai-liu2024bootstrapping-r2-llm}, LlamaRG \cite{2026-CVPR-LlamaRG}, Libra \cite{acl-2025-libra}, and CheXagent \cite{chen2024-chexagent}. Other work uses reinforcement learning to optimize clinical and linguistic quality \cite{xiao2025radiology-mpo-aaai-2025}, or adapts general-purpose and domain-specific VLMs, including LLaVA-Med \cite{NEURIPS2023-llava-med}, MambaXray-VL-L \cite{Wang-2025-CVPR-CXPMRG-Bench}, our MLRG \cite{Liu-2025-CVPR-mlrg}, and R2GenAlign \cite{2026-CVPR-r2genalign}. CheXLoc complements these efforts by learning visual representations through entity-guided local patch--token alignment.

\section{Methods}
\label{sec:method}

\subsection{Architecture of CheXLoc}
CheXLoc learns locally aligned visual representations from paired chest radiographs and reports without spatial annotations. It is based on the premise that report tokens provide different levels of supervision: clinical entities describing radiographic findings are more directly related to local visual evidence than function words and syntactic fillers. Conventional alignment strategies \cite{wang-mgca,2026-nbe-afloc} that weight all tokens equally may therefore dilute clinically informative supervision.

Following weakly supervised localization~\cite{2022-pami-weakly-supervised-object}, CheXLoc uses report-derived clinical entities as coarse semantic supervision for local visual--textual correspondence. We formulate entity--patch learning as a multiple-instance learning (MIL) problem~\cite{2018-icml-multiple-instance-learning}: each radiograph is treated as a bag of patches, only a subset of which is expected to contain evidence relevant to a given entity. As shown in Fig.~\ref{fig:overview}c, CheXLoc combines (i) global image--report alignment to capture holistic correspondence and (ii) entity-guided local patch--token alignment, which estimates token importance from clinical entities and uses these weights to modulate local alignment. Token-level supervision is derived from paired reports using RadGraph~\cite{jain-radgraph} and a manually curated clinical lexicon; no spatial labels are used during pretraining.

\textbf{Text Encoding.}
We employ a shared language encoder to obtain representations of radiology reports and associated clinical context. To distinguish among clinical sections, section-specific tokens ([Findings], [Indication], [History], [Comparison], and [Technique]) are prepended to the corresponding input sequences. The augmented textual inputs are encoded using CXR-BERT~\cite{2022-eccv-cxr-bert}. The resulting representations are projected into a shared vision--language embedding space of dimension $d$ using multilayer perceptron (MLP)-based projection heads. We denote the resulting report and contextual representations as $\boldsymbol{T}^{r}\in\mathbb{R}^{N^{r}\times d}$ and $\boldsymbol{T}^{c}\in\mathbb{R}^{N^{c}\times d}$, where $N^{r}$ and $N^{c}$ represent the corresponding sequence lengths.

\textbf{Image Encoding.}
Following our PriorRG~\cite{PriorRG}, we use RADDINO~\cite{2024-rad-dino-nmi} to extract patch-level features from the current radiograph and, when available, the most recent prior examination. If no prior examination is available, only the current image is used. An MLP projects the visual features into the shared embedding space, and a Vision Transformer block~\cite{2021-vit} models interactions among tokens across time points. The resulting representation is denoted by $\boldsymbol{X}\in\mathbb{R}^{N^{v}\times d}$, where $N^{v}$ is the number of image patches.

\textbf{Global Image--Report Alignment.}
Radiological interpretation depends on clinical context, including symptoms, examination indications, prior history, and imaging techniques. We therefore adopt the context-guided fusion module from PriorRG~\cite{PriorRG}. A Perceiver module~\cite{icml-2021-perceiver} with learnable latent queries aggregates the contextual representations $\boldsymbol{T}^{c}$ and fuses them with visual features $\boldsymbol{X}$. Global image--report correspondence is then optimized with the multi-positive contrastive objective from PriorRG~\cite{PriorRG}, denoted by $\mathcal{L}_{\mathrm{global}}$, which pulls matched pairs together and separates unmatched pairs within a mini-batch.

\textbf{Entity-Informed Token Importance Estimation.}
Existing token-level alignment methods~\cite{wang-mgca,2026-nbe-afloc} typically assign equal weights to textual tokens when establishing local visual--textual correspondence. However, the relevance of report tokens to local visual evidence varies with their clinical semantics. Clinical entities, such as \textit{opacity}, \textit{effusion}, and \textit{cardiomegaly}, provide more direct supervision for local alignment than function words and syntactic fillers. Uniform token weighting may therefore dilute the contribution of clinically relevant information.

To address this issue, CheXLoc derives token weights from report-level clinical entities and their semantic categories, rather than relying solely on token occurrence frequency~\cite{zhao2025-superclip}. We use RadGraph~\cite{jain-radgraph} to extract entities and their corresponding types. Since the entity extractor may omit clinically relevant expressions, such as \textit{no} and \textit{stable}, we supplement the extracted entities with a manually curated lexicon that includes negation and temporal descriptors. For the $i^{th}$ report, the extracted entity--type pairs are represented as:
\begin{equation}
	\{(e_{i,j},r_{i,j})\}_{j=1}^{N_i^e}, r_{i,j} \in \mathcal{T},
\end{equation}
where $e_{i,j}$ denotes the $j^{th}$ entity span, $r_{i,j}$ indicates its entity type, $\mathcal{T}$ is the set of entity types, and $N_i^e$ represents the total number of extracted entities in the $i^{th}$ report. To model the varying contributions of different entity types to local visual--textual alignment, we introduce a learnable importance function $\boldsymbol{\alpha}: \mathcal{T} \to \mathbb{R}_{+}$. It assigns an importance weight to each entity type, with $\boldsymbol{\alpha}(r_{i,j})$ denoting the weight of the $j^{th}$ entity. The function $\boldsymbol{\alpha}(\cdot)$ is optimized jointly with the pretraining objective, allowing the relative importance of entity types to be adapted during training.

Since local alignment is performed at the token level, we map the learned entity-type weights to the tokenizer tokens within each entity span. Specifically, the importance weight of each vocabulary token $k$ is obtained by aggregating its occurrences across across entity spans, weighted by their corresponding entity-type importance scores:
%Because vision--language pretraining operates on tokens, we assign each entity's importance to the tokenizer tokens within its span. This maps entity-level supervision to token-level weights while retaining compatibility with autoregressive report generation. For each vocabulary token $k$, its importance is aggregated over all occurrences in the report:
\begin{equation}
	\boldsymbol{w}_{i,k} = \sum_{j=1}^{N_i^e} \boldsymbol{\alpha}(r_{i,j}) \cdot {\text{Freq}}(k, e_{i,j}), k \in\mathcal{V},
\end{equation}
where $\mathrm{Freq}(k,e_{i,j})$ denotes the number of occurrences of token $k$ within entity span $e_{i,j}$, and $\mathcal{V}$ represents the tokenizer vocabulary. The resulting $\boldsymbol{w}_{i,k}$ represents the entity-informed importance weight of token $k$ in the $i^{th}$ report, obtained by aggregating the importance scores of the entity types associated with its occurrences.

\textbf{Entity-Guided Local Patch--Token Alignment Objective.}
The entity-informed token weights are incorporated into local patch--token alignment to account for the varying contributions of report tokens. Specifically, patch-level visual features $\boldsymbol{X}$ are used to predict token-level clinical semantics. To further account for the difficulty of visual prediction, we introduce an adaptive reweighting factor $\boldsymbol{\beta}$, which jointly modulates the contributions of positive and negative tokens to the alignment objective. Specifically, we first derive binary token-level targets from the estimated token importance scores:
\begin{equation}
	\boldsymbol{b} = \mathbb{I}(\boldsymbol{w} > 0) \in {\mathbb R}^{B \times |\mathcal{V}|},
\end{equation}
where $\mathbb{I}(\cdot)$ denotes the indicator function, $B$ is the batch size, and $|\mathcal{V}|$ represents the vocabulary size. Thus, $\boldsymbol{b}_{i,j}=1$ indicates the $j^{th}$ token is part of a clinical entity in the $i^{th}$ sample. To account for both clinical importance and prediction difficulty, we introduce an adaptive reweighting factor $\boldsymbol{\beta} \in {\mathbb R}^{B \times |\mathcal{V}|}$. Unlike a probability distribution, $\boldsymbol{\beta}$ serves as an optimization weight that modulates the contribution of individual tokens to the training objective. We first transform the token importance scores as follows:
\begin{align} 
	\tilde{\boldsymbol{w}} &= \log(1+\boldsymbol{w}),\\
	\boldsymbol{\beta} &= \boldsymbol{b} \odot \left [ \boldsymbol{p} + (1-\boldsymbol{p}) \odot \tilde{\boldsymbol{w}} \right ] + (1-\boldsymbol{b}) \odot  \boldsymbol{p}.
\end{align}
Here, $\odot$ denotes the Hadamard product, and $\boldsymbol{p}\in{\mathbb R}^{B\times |\mathcal{V}|}$ represents token probabilities predicted from patch-level visual representations $\boldsymbol{X}$. The probability $\boldsymbol{p}$ estimates prediction confidence, whereas $\tilde{\boldsymbol{w}}$ encodes clinical importance. Their integration in $\boldsymbol{\beta}$ adaptively modulates the contribution of each token to the optimization objective. For a positive token ($\boldsymbol{b}_{i,j}=1$), the corresponding weight is given by:
\begin{equation}
	\boldsymbol{\beta}_{i,j}=\boldsymbol{p}_{i,j}+(1-\boldsymbol{p}_{i,j}) \odot \tilde{\boldsymbol{w}}_{i,j}.
\end{equation}
Thus, positive tokens with low predicted probabilities ($\boldsymbol{p}_{i,j} \rightarrow 0$) receive larger weights, emphasizing hard positives. As the predicted probability of a positive token approaches one ($\boldsymbol{p}_{i,j} \rightarrow 1$), its weight approaches one.  For a negative token ($\boldsymbol{b}_{i,j}=0$), the weight becomes $\boldsymbol{\beta}_{i,j}=\boldsymbol{p}_{i,j}$, emphasizing hard negatives and downweighting easy negatives.  To further enhance optimization stability, we incorporate focal modulation~\cite{2017-iccv-focal-loss}:
%are treated as hard positives and receive larger optimization weights, thereby inducing stronger gradients. Conversely, as a token becomes increasingly well captured by the visual representations ($\boldsymbol{p}_{i,j} \rightarrow 1$), its weighting factor approaches unity, thereby avoiding excessive optimization of tokens that are already well represented. For a negative token ($\boldsymbol{b}_{i,j}=0$), its weighting factor becomes $\boldsymbol{\beta}_{i,j}=\boldsymbol{p}_{i,j}$. This formulation assigns greater weights to hard negatives with high predicted probabilities while downweighting easy negatives with low predicted probabilities. Such adaptive reweighting enables CheXLoc to focus optimization on clinically informative and visually challenging tokens. To further enhance optimization stability, we incorporate focal modulation~\cite{2017-iccv-focal-loss}:
\begin{equation} 
	\label{eq:reweight}
	{\cal L}_{i,j} = \boldsymbol{\beta}_{i,j} \cdot (1 -  \tilde{\boldsymbol{p}}_{i,j})^{\gamma} \cdot \text{BCE}(\boldsymbol{p}_{i,j},\boldsymbol{b}_{i,j}),
\end{equation}
where $\tilde{\boldsymbol{p}}_{i,j} = \boldsymbol{b}_{i,j} \cdot \boldsymbol{p}_{i,j} + (1 - \boldsymbol{b}_{i,j}) \cdot (1-\boldsymbol{p}_{i,j})$, $\gamma=2$ controls the focal modulation strength, and $\text{BCE}(\cdot)$ is the binary cross-entropy loss. To prevent optimization from being dominated by numerous uninformative negative tokens, the loss is normalized by the effective number of informative tokens:
\begin{equation} 
	N_i = \max\left \{ 1,  {\textstyle \sum_{j}^{}} \left [ \boldsymbol{b}_{i,j} + (1-\boldsymbol{b}_{i,j}) \cdot \mathbb{I}(\boldsymbol{p}_{i,j} > \tau ) \right ]   \right \},
\end{equation}
where $\tau=0.1$ controls the selection of hard negatives. The final token-level alignment objective is:
\begin{equation} 
	{\cal L}_{\text{local}} = \frac{1}{B}\sum_{i=1}^{B} \frac{1}{N_i} \sum_{j=1}^{|\mathcal{V}|}{\cal L}_{i,j} .
\end{equation}
This objective limits the contribution of abundant easy negatives while maintaining emphasis on clinically relevant tokens and difficult visual--textual associations.

\textbf{Training Objective.}
The overall pretraining objective jointly optimizes global image--report alignment and entity-guided local patch--token alignment:
\begin{align}
	{\cal L}_{\text{pretrain}}={\cal L}_{\text{global}}+{\cal L}_{\text{local}}.
\end{align}
This objective encourages locally aligned visual representations to emerge from image--report pairs, without requiring spatial annotations.

\subsection{Localization-Aware Evaluation for Report Generation} 
\label{metrics}
Existing explanation metrics, including sufficiency and comprehensiveness~\cite{2020-acl-ERASER}, insertion and deletion~\cite{Petsiuk2018rise}, and smallest sufficient regions~\cite{nips-2017-smallest-sufficient-regions}, were developed mainly for classification and post hoc explanation. They do not directly measure how generated findings respond to disease-specific image regions. We therefore introduce Localization-Aware Generation Evaluation (LAGE), a perturbation-based framework that measures the sensitivity of generated findings to lesion-specific image evidence.

For a target lesion $c$, let $x$ denote the original chest radiograph and $M_c$ the corresponding lesion mask. We construct two complementary perturbations. A lesion-only image retains only the target lesion region,
\begin{align}
	x_c^{\mathrm{only} } = x \odot M_c.
\end{align}
whereas a lesion-removed image removes the target lesion while preserving all remaining image content,
\begin{align}
	x_c^{\mathrm{rm} } = x \odot (1-M_c).
\end{align}
Following the CheXbert evaluation protocol \cite{Smit2020_chexbert}, each generated report is converted into a binary vector of clinical observations:
\begin{align}
	\hat{\boldsymbol{y}}_i = \left ( \hat{y}_{i,1}, \dots, \hat{y}_{i,K} \right ) \in \left \{ 0,1 \right \} ^K,
\end{align}
where $K$ denotes the number of clinical observations, and ${\boldsymbol{y}_i} = \left ( {y}_{i,1}, \dots, {y}_{i,K} \right ) \in \left \{ 0,1 \right \} ^K$ indicates the corresponding reference labels derived from the ground-truth report.

\textbf{(1) Sufficiency.} Sufficiency measures whether the isolated lesion provides evidence for the target finding while suppressing unrelated findings. For a target lesion $c$, the sufficiency score for the $i^{th}$ sample is defined as:
\begin{align}
	S_i(c) = \mathbb{I}\left ( \hat{y}_{i,c}^{\mathrm{only} } = y_{i,c}  \right ) \left ( 1-\frac{1}{K-1}\sum_{k\ne c}^{}  \hat{y}_{i,k}^{\mathrm{only} }  \right ),
\end{align}
where $\hat{y}_{i,c}^{\mathrm{only}}$ denotes the predicted label of lesion $c$ for the $i^{th}$ sample computed from the lesion-only image. The metric is intentionally designed to reward both recovery of the target finding and suppression of unrelated findings under lesion-only input. The sufficiency for lesion $c$ is computed as:
\begin{align}
	\mathrm{Sufficiency}(c) = \frac{1}{N_c} \sum_{i=1}^{N_c} S_i(c),
\end{align}
where $N_c$ is the number of samples containing lesion $c$. A high sufficiency score indicates that the lesion-only image supports the target finding while limiting additional predictions. This definition is conservative: it penalizes every additional predicted finding, including findings present in the original report.

% =================comparion of existing methods in pretraining set==============
\begin{table}
	\centering
	\caption{Comparison of existing methods in terms of the number of parameters, pretraining dataset scale, and supervision type.}
	\label{tab:baselines-pretraining-dataset}
	\setlength{\tabcolsep}{0.65mm}
	\begin{tabular}{ccccc} 
		\toprule
		\textbf{Method} & \textbf{\#Params} & \textbf{\#Images} & \textbf{Supervision} & \textbf{Data sources} \\ 
		\midrule
		AFLoc & 231M & 220K & image-report & MIMIC-CXR+others \\
		RADDINO & 183M & 838,336 & image-only & MIMIC-CXR+others \\
		CheXWorld & 182M & $\sim$500K & image-only & MIMIC-CXR+others \\
		BiomedCLIP & 292M & $\sim$15.28M & image-report & PMC-15M \\
		EGMA & 108M & 3,689 & image-report+Spat. & MIMIC-CXR+others \\
		Libra & 3B & $\sim$1.22M & image-report & MIMIC-CXR+others \\
		CheXagent & 3B & $\sim$1.05M & image-report & MIMIC-CXR+others \\
		MedGemma & 4B & $\sim$33.14M & image-report & MIMIC-CXR+others \\
		MedVersa & 7B & $\sim$29M & image-report+Spat. & MIMIC-CXR+others \\
		MAIRA-2 & 7B & 501,825 & image-report+Spat. & MIMIC-CXR+others \\
		RadFM & 14B & $\sim$16M & image-report & MedMD \\
		\textbf{CheXLoc} & 282M & 240,197 & image-report & MIMIC-CXR \\
		\bottomrule
	\end{tabular}
	
	\begin{minipage}{\linewidth}
		\vspace{1mm}
		\footnotesize
		\textit{Notes:} Spat. denotes spatial annotations.
	\end{minipage}
\end{table}

\textbf{(2) Necessity.} Necessity evaluates whether removing a lesion suppresses the target finding while leaving other findings unchanged. Following lesion removal, the label corresponding to the target finding is set to zero, while all other labels are retained. This counterfactual labeling assumes that removing the annotated target lesion eliminates its corresponding clinical finding while leaving other findings unchanged, solely for the purpose of evaluation and without implying clinical causality. Based on this assumption, the counterfactual reference labels are defined as:
\begin{align}
	{\tilde{\boldsymbol y}}_i = \left ( {y}_{i,1}, \dots, 0, \dots, {y}_{i,K} \right ).
\end{align}
Necessity is computed as the macro-F1 score between predictions $\hat{\boldsymbol y}_i^{\mathrm{rm} }$ generated from the lesion-removed image and the modified reference labels, denoted as:
\begin{align}
	\mathrm{Necessity}(c) = \frac{1}{N_c} \sum_{i=1}^{N_c} \text{Macro-F1}\left ( \hat{\boldsymbol y}_i^{\mathrm{rm} },  \tilde{\boldsymbol y}_i \right )  ,
\end{align}

%A high necessity score indicates that lesion removal suppresses the target clinical finding without affecting unrelated findings, demonstrating that the removed lesion is necessary for generating the corresponding clinical finding.

\textbf{(3) Localization-Aware Generation Evaluation (LAGE).} To jointly evaluate these complementary properties, we define Localization-Aware Generation Evaluation (LAGE) as the arithmetic mean of sufficiency and necessity:
\begin{align}
	\mathrm{LAGE}(c) = \frac{1}{2}\left ( \mathrm{Sufficiency}(c)+\mathrm{Necessity}(c) \right ).
\end{align}
LAGE measures the sensitivity of generated findings to controlled lesion-specific perturbations. It provides evidence of localization awareness but does not establish causal image--text grounding.

\subsection{Ethical Considerations}
This study involved secondary analyses of de-identified chest X-ray datasets, including MIMIC-CXR, ReXGradient, CheXpert Plus, IU X-ray, COVIDx, TBX11K, SIIM, RSNA, and Shenzhen. The MIMIC-CXR dataset was obtained from PhysioNet through its credentialed access process, which required completion of the CITI Data or Specimens Only Research training, PhysioNet credentialing, and signing of the corresponding Data Use Agreement (DUA). %According to the original MIMIC-CXR study, the data collection and resource development were approved by the Institutional Review Board (IRB) of Beth Israel Deaconess Medical Center (BIDMC), with individual informed consent waived. 
The other datasets were used in accordance with their respective data access policies and ethical requirements. No new human participants were recruited, and no additional patient data were collected in this study.

% ===================report generation results on MVLM
\begin{figure*}
	\centering
	\includegraphics[width=1\linewidth]{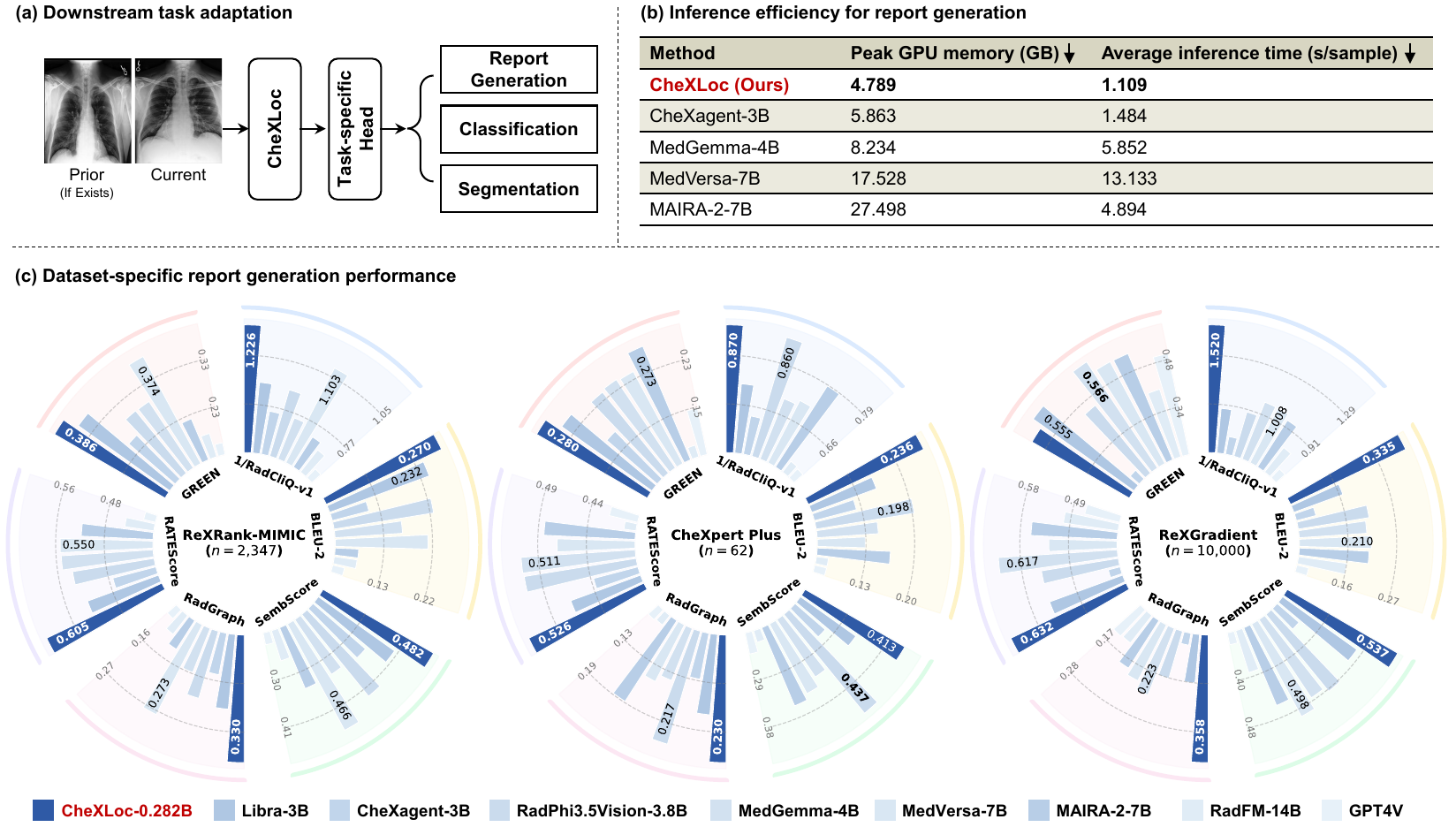}
	\caption{
		\textbf{Evaluation of chest X-ray report generation performance and inference efficiency.}
		\textbf{a,} Downstream task adaptation, in which current and, when available, prior chest X-ray images are encoded into unified visual representations and processed by task-specific prediction heads.
		\textbf{b,} Peak GPU memory consumption and average inference time per sample for report generation on the ReXRank-MIMIC \cite{zhang2025rexrank} test subset ($n = 2{,}347$).
		\textbf{c,} Quantitative comparison of report generation performance across ReXRank-MIMIC ($n = 2{,}347$), CheXpert Plus ($n = 62$), and ReXGradient ($n = 10{,}000$) using 1/RadCliQ-V1, BLEU-2, SembScore, RadGraph, RATEScore, and GREEN.
	}
	\label{fig:report-generation}
\end{figure*}

% ========================inference time======================
\begin{table}
	\caption{Inference efficiency of CheXLoc across batch sizes for report generation on ReXRank-MIMIC dataset ($n=2{,}347$).}
	\label{tab:supp_inference_time}
	\centering
	\begin{tabular}{ccc} 
		\toprule
		\textbf{Batch size} & \textbf{Peak GPU memory} & \textbf{Average inference time $\downarrow$} \\ 
		\midrule
		1 & 4.789 GB & 1.109 s/sample \\
		2 & 5.092 GB & 0.662 s/sample \\
		4 & 5.698 GB & 0.416 s/sample \\
		8 & 6.912 GB & 0.273 s/sample \\
		16 & 9.337 GB & 0.246 s/sample \\
		32 & 14.187 GB & 0.263 s/sample \\
		64 & 23.163 GB & 0.264 s/sample \\
		\bottomrule
	\end{tabular}
\end{table}

\section{Experiments}
\subsection{Experimental Settings}
\textbf{Pretraining Dataset.} We pretrain CheXLoc on the official MIMIC-CXR training split~\cite{johnson-mimic-cxr-jpg}, which contains 59,693 patients, 240,197 radiographs, and 150,957 reports. For each downstream task, the corresponding test set is strictly excluded from pretraining and adaptation. The pretraining data used by the baselines are summarized in Tab.~\ref{tab:baselines-pretraining-dataset}.

\textbf{Report Generation Datasets.} For MIMIC-CXR \cite{johnson-mimic-cxr-jpg}, we use the official training, validation, and test splits of 240,197, 2,113, and 3,852 radiographs, respectively. We also report results on the ReXRank-MIMIC subset ($n=2{,}347$)~\cite{zhang2025rexrank}, drawn from the MIMIC-CXR test set. For CheXpert Plus \cite{chexpert-plus-2024}, we follow the ReXRank protocol, using 27,299 patients for training and 62 for testing. ReXGradient \cite{2025-ml4h-rexgradient} is evaluated using its official split of 140,000/10,000/10,000 studies for training/validation/testing. The full IU X-ray dataset ($n=2{,}955$)~\cite{demner2016preparing-iu-xray} is reserved for zero-shot evaluation.

\textbf{Classification and Segmentation Datasets.} We use COVIDx \cite{2022-covidx} for COVID-19, Shenzhen \cite{2023-shenzhencxr} and TBX11K \cite{tbx11k-cvpr-2020} for tuberculosis, SIIM \cite{siim-acr-pneumothorax-segmentation} for pneumothorax, and RSNA \cite{shih2019augmenting-rsna} for pneumonia. COVIDx contributes 400 cases to zero-shot testing. Following CheXWorld \cite{2025-cvpr-chexworld}, we split Shenzhen into 463/65/134 images for training/validation/testing. Following BenchX \cite{zhou2024benchx}, the corresponding splits for SIIM, TBX11K, and RSNA contain 9,303/1,372/1,372, 5,879/1,260/1,260, and 18,678/4,003/4,003 images, respectively.

\textbf{Implementation Details.}\label{sec:inference}
CheXLoc is pretrained on the MIMIC-CXR training split. The architecture, image preprocessing, data augmentation, and optimization settings follow our PriorRG~\cite{PriorRG}. Radiographs are resized to $518\times518$ pixels. We optimize the model with AdamW, using an initial learning rate of $5\times10^{-5}$ and a \texttt{ReduceLROnPlateau} scheduler with a patience of two epochs. Pretraining runs for 10 epochs with a batch size of 48 on one NVIDIA RTX 5880 Ada GPU and takes approximately 30 h. The initial importance weights for anatomy, observation, negation, and temporal entities in $\boldsymbol{\alpha}$ are 1.5, 2.0, 2.0, and 1.5, respectively.

We adapt the pretrained encoders to report generation, disease classification, and lesion segmentation (Fig.~\ref{fig:report-generation}a). For methods without an original decoder, we use DistilGPT-2~\cite{Sanh2019DistilBERTAD} and jointly optimize the encoder and decoder. Zero-shot report generation on IU X-ray uses models fine-tuned on MIMIC-CXR. Disease classification is evaluated in zero-shot and supervised settings following the CLIP paradigm~\cite{radford-learning-clip}, with shared prompts and training subsets across methods. For segmentation, we attach a lightweight convolutional decoder to each pretrained visual encoder and jointly optimize the resulting model using Dice and focal losses. Task-specific learning rates follow their respective official implementations.

Inference efficiency is evaluated on the ReXRank-MIMIC test subset ($n=2{,}347$) under identical hardware conditions. For fair comparison, all methods are evaluated with a batch size of 1 and their official inference settings (Fig.~\ref{fig:report-generation}b). Average inference time per sample is measured as wall-clock time, including preprocessing and report generation, while peak GPU memory is recorded. CheXLoc also supports batched inference, with the corresponding trade-off between inference time and GPU memory summarized in Tab.~\ref{tab:supp_inference_time}. Increasing the batch size to 16 reduces the average inference time from 1.109 to 0.246~s/sample, while increasing GPU memory usage from 4.789 to 9.337~GB. Larger batch sizes provide negligible additional speedup at substantially higher memory costs.

% ===================in-domain report generation on mimic-cxr (n=3852)========================
\begin{table*}
	\centering
	\caption{report generation performance on the MIMIC-CXR dataset \cite{johnson-mimic-cxr-jpg}. EGMA is reproduced using its official code with DistilGPT2, while other results are cited from original papers ({\color{red}{\textbf{Best}}}, {\color{blue}{{{Second Best}}}}).}
	\setlength{\tabcolsep}{1.3mm}
	\label{tab:rg-mimic-cxr}
	\begin{tabular}{ccccccccccc} 
		\toprule
		\textbf{Model} & \textbf{Venue} & \textbf{BLEU-1$\uparrow$} & \textbf{BLEU-2$\uparrow$} & \textbf{BLEU-3$\uparrow$} & \textbf{BLEU-4$\uparrow$} & \textbf{METEOR$\uparrow$} & \textbf{ROUGE-L$\uparrow$} & \textbf{Precision$\uparrow$} & \textbf{Recall$\uparrow$} & \textbf{$^{14}$Mi-F1$\uparrow$ } \\
		MAN \cite{shen2024automatic_aaai} & AAAI'24 & 0.396 & 0.244 & 0.162 & 0.115 & 0.151 & 0.274 & 0.411 & 0.398 & 0.389 \\
		R2-LLM \cite{aaai-liu2024bootstrapping-r2-llm} & AAAI'24 & 0.402 & 0.262 & 0.180 & 0.128 & 0.175 & 0.291 & 0.465 & 0.482 & 0.473 \\
		SEI \cite{sei} & MICCAI'24 & 0.382 & 0.247 & 0.177 & 0.135 & 0.158 & 0.299 & 0.523 & 0.410 & 0.460 \\
		EGMA \cite{ma2024-nips-eye-egma}  & NeurIPS'24 & 0.395 & 0.260 & 0.183 & 0.132 & 0.184 & 0.307 & 0.500 & 0.453 & 0.475 \\
		%HERGen \cite{2024-eccv-hergen} & ECCV'24 & 0.395 & 0.248 & 0.169 & 0.122 & 0.156 & 0.285 & - & - & - \\
		PromptMRG \cite{promtmrg-aaai-2024} & AAAI'24 & 0.398 & -  & - & 0.112 & 0.157 & 0.268 & 0.501 & \color{red}{\textbf{0.509}} & 0.476 \\
		MPO \cite{xiao2025radiology-mpo-aaai-2025} & AAAI'25 & 0.416 & 0.269 & 0.191 & 0.139 & 0.162 & 0.309 & 0.436 & 0.376 & 0.353 \\
		MambaXray-VL-L \cite{Wang-2025-CVPR-CXPMRG-Bench} & CVPR'25 & 0.422 & 0.268 & 0.184 & 0.133 & 0.167 & 0.289 &  & 0.460 & 0.505 \\
		MLRG \cite{Liu-2025-CVPR-mlrg} & CVPR'25 & 0.411 & 0.277 & 0.204 & 0.158 & 0.176 & 0.320 & 0.549 & 0.468 & 0.505 \\
		SAT-RRG \cite{sat-rrg} & CVPR'26 & 0.428 & 0.280 & 0.196 & 0.143 & 0.167 & 0.303 & - & - & - \\
		R2GenAlign \cite{2026-CVPR-r2genalign} & CVPRF'26 & \color{blue}{0.433} & 0.285 & 0.204 & 0.145 & 0.175 & 0.318 & - & - & - \\
		LlamaRG \cite{2026-CVPR-LlamaRG} & CVPRF'26 & 0.417 & 0.270 & 0.178 & 0.130 & 0.162 & 0.302 & 0.448 & 0.431 & 0.414 \\
		PriorRG \cite{PriorRG} & AAAI'26 & 0.412 & \color{blue}{0.290} & \color{blue}{0.220} & \color{blue}{0.175} & \color{blue}{0.189} & \color{blue}{0.324} & 0.541 & 0.485 & \color{blue}{0.511} \\
		MATM \cite{MATM-2025-tpami} & TPAMI'25 & 0.387 & 0.251 & 0.177 & 0.129 & 0.164 & 0.291 & - & - & - \\
		PCVs \cite{pcvs-26-tpami} & TPAMI'26 & - & - & - & 0.113 & 0.162 & 0.317 & 0.440 & 0.475 & 0.442 \\
		AM-MRG \cite{AM-MRG-2026-tmi} & TMI'26 & 0.426 & 0.238 & 0.159 & 0.109 & 0.173 & 0.282 & \color{blue}{0.555} & 0.429 & 0.484 \\
		\textbf{CheXLoc} & Ours & \color{red}{\textbf{0.449}} & \color{red}{\textbf{0.340}} & \color{red}{\textbf{0.273}} & \color{red}{\textbf{0.226}} & \color{red}{\textbf{0.217}} & \color{red}{\textbf{0.380}} & \color{red}{\textbf{0.571}} & \color{blue}{{0.502}} & \color{red}{\textbf{0.534}} \\
		\bottomrule
	\end{tabular}
\end{table*}

% ==================zero-shot iu-xray performance ====================
\begin{table}
	\centering
	\caption{Zero-shot report generation performance on the external IU X-ray dataset without fine-tuning ({{\textcolor{red}{\textbf{Best}}}}, {\textcolor{blue}{Second Best}}).}
	\label{tab:rg-iu-xray}
	\setlength{\tabcolsep}{0.50mm}
	\begin{tabular}{cccccc} 
		\toprule
		\textbf{Model} & \textbf{RadGraph$\uparrow$} & \textbf{SembScore$\uparrow$} & \textbf{1/RQ$\uparrow$} & \textbf{RATEScore$\uparrow$} & \textbf{BLEU-2$\uparrow$} \\ 
		\midrule
		MAIRA-2-7B & 0.193 & 0.503 & 0.881 & 0.543 & 0.170 \\
		MedGemma-4B & \textcolor{red}{\textbf{0.294}} & \textcolor{blue}{0.559} & \textcolor{red}{\textbf{1.132}} & \textcolor{red}{\textbf{0.634}} & \textcolor{red}{\textbf{0.208}} \\
		MedVersa-7B & 0.008 & 0.313 & 0.505 & 0.455 & 0.000 \\
		CheXagent-3B & 0.235 & 0.492 & 0.899 & 0.552 & 0.160 \\
		AFLoc & 0.230 & 0.524 & 0.976 & 0.595 & 0.180 \\
		CheXWorld & 0.206 & 0.540 & 0.969 & 0.587 & 0.188 \\
		BiomedCLIP & 0.228 & 0.473 & 0.911 & 0.585 & 0.180 \\
		RADDINO & 0.198 & 0.547 & 0.973 & 0.592 & 0.195 \\
		EGMA & 0.229 & 0.486 & 0.796 & 0.539 & 0.093 \\
		\textbf{CheXLoc} & \textcolor{blue}{0.240} & \textcolor{red}{\textbf{0.565}} & \textcolor{blue}{1.073} & \textcolor{blue}{0.609} & \textcolor{blue}{0.193} \\
		\bottomrule
	\end{tabular}
	
	\begin{minipage}{\linewidth}
		\vspace{1mm}
		\footnotesize
		\textit{Notes:} 1/RQ denotes 1/RadCliQ-V1.
	\end{minipage}
	
\end{table}

\subsection{Evaluation Metrics}
\textbf{Chest X-ray Report Generation.} Following the ReXRank \cite{zhang2025rexrank} protocol, report generation performance is assessed using both lexical and clinically oriented metrics. BLEU-n measures n-gram overlap between generated and reference reports. Clinical performance is evaluated using SembScore \cite{Smit2020_chexbert}, RadGraph \cite{jain-radgraph}, RATEScore \cite{zhao2024ratescore}, GREEN \cite{2024-green}, 1/RadCliQ-V1 \cite{yu2023evaluating}, $^n$Ma-F1, and $^n$Mi-F1. Specifically, SembScore evaluates clinical semantic similarity using pathology representations derived from CheXbert \cite{Smit2020_chexbert}, RadGraph assesses agreement in clinical entities and relations, and RATEScore and GREEN evaluate clinical report quality and clinically significant errors, respectively. RadCliQ-V1 \cite{yu2023evaluating} is a composite metric that predicts the number of radiologist-assessed clinical reporting errors using BLEU-2, BERTScore, CheXbert vector similarity, and RadGraph. $^n$Ma-F1 and $^n$Mi-F1 denote the macro- and micro-F1 scores, respectively, computed by CheXbert \cite{Smit2020_chexbert} over $n\in\{5,14\}$ observations. The five-observation setting includes \textit{Cardiomegaly}, \textit{Edema}, \textit{Consolidation}, \textit{Atelectasis}, and \textit{Pleural Effusion}. Following ReXRank~\cite{zhang2025rexrank}, we report $1/\mathrm{RadCliQ\text{-}V1}$. Higher scores indicate better performance for all metrics.

\textbf{Disease Classification and Lesion Segmentation.} Disease classification performance is measured using the area under the receiver operating characteristic curve (AUROC), while lesion segmentation performance is evaluated using the mean Dice coefficient to quantify the overlap between predicted and reference masks.

\textbf{Statistical Analysis.} We compare CheXLoc with each baseline using 1,000 paired bootstrap resamples within each evaluation setting. The 95\% confidence intervals (CIs) are given by the 2.5th and 97.5th percentiles of the bootstrap distributions. Two-sided $p$ values are computed from the centered bootstrap distribution of paired performance differences. %Statistical significance is assessed at nominal thresholds of $p < 0.05$, $p < 0.01$, and $p < 0.001$.

\subsection{Report-Supervised Pretraining Supports Efficient and Transferable Chest X-ray Report Generation}

We evaluate whether report-supervised pretraining yields effective and transferable visual representations for chest X-ray report generation. Experiments cover in-domain evaluation on MIMIC-CXR and ReXRank-MIMIC, cross-dataset evaluation on CheXpert Plus and ReXGradient, and zero-shot transfer to an independent institution.

\textbf{Efficient Report Generation on In-Domain Chest Radiographs.}
We first evaluate CheXLoc on the ReXRank-MIMIC ($n = 2{,}347$), enabling direct comparison with the publicly reported benchmark results. As shown in Fig.~\ref{fig:report-generation}c, CheXLoc achieves the highest numerical performance across all evaluated metrics, including 1/RadCliQ-V1 \cite{yu2023evaluating}, BLEU-2, SembScore \cite{Smit2020_chexbert}, RadGraph \cite{jain-radgraph}, RATEScore \cite{zhao2024ratescore}, and GREEN \cite{2024-green}. The difference is particularly pronounced for clinically oriented metrics: CheXLoc achieves a 1/RadCliQ-V1 score of 1.226, compared with 0.898 for Libra-3B \cite{acl-2025-libra}, 0.694 for MAIRA-2-7B \cite{bannur2024maira2groundedradiologyreport}, and 0.650 for RadFM-14B \cite{wu2025towards-radfm}. The RadGraph and GREEN results also indicate stronger agreement in extracted clinical content and fewer clinically significant errors. Importantly, CheXLoc achieves these results with 0.282 billion parameters, fewer than the compared multimodal foundation models.

We also compare CheXLoc with existing chest X-ray report generation methods on the MIMIC-CXR test set ($n=3{,}852$; Tab.~\ref{tab:rg-mimic-cxr}). CheXLoc achieves the highest scores on most metrics, exceeding the second-best methods by 5.1 and 2.3 percentage points in BLEU-4 and $^{14}$Mi-F1, respectively. These results support the effectiveness of the learned visual representations for chest X-ray report generation.

%We further evaluated CheXLoc against medical vision--language models \cite{2026-nbe-afloc,2024-rad-dino-nmi,2025-cvpr-chexworld,zhang2025biomedclipmultimodalbiomedicalfoundation,ma2024-nips-eye-egma} on the full MIMIC-CXR test set (Table~\ref{tab:rg-medvlm}; $n = 3{,}852$). CheXLoc achieved the highest scores across all evaluated metrics (all nominal $p<0.001$). Its RadGraph score reached 0.329, compared with 0.205 for the highest-scoring competing model, while SembScore, 1/RadCliQ-V1, and RATEScore reached 0.482, 1.303, and 0.604, respectively. The BLEU score also reached 0.340, consistent with improvements in both clinical semantic similarity and lexical correspondence with the reference reports.
%Collectively, these results indicate that the localized visual representations learned by CheXLoc provide clinically informative image--report alignment, contributing to higher agreement between generated and reference reports across the evaluated metrics.

% ========================segmentation===================================
\begin{figure*}
	\centering
	\includegraphics[width=1\linewidth]{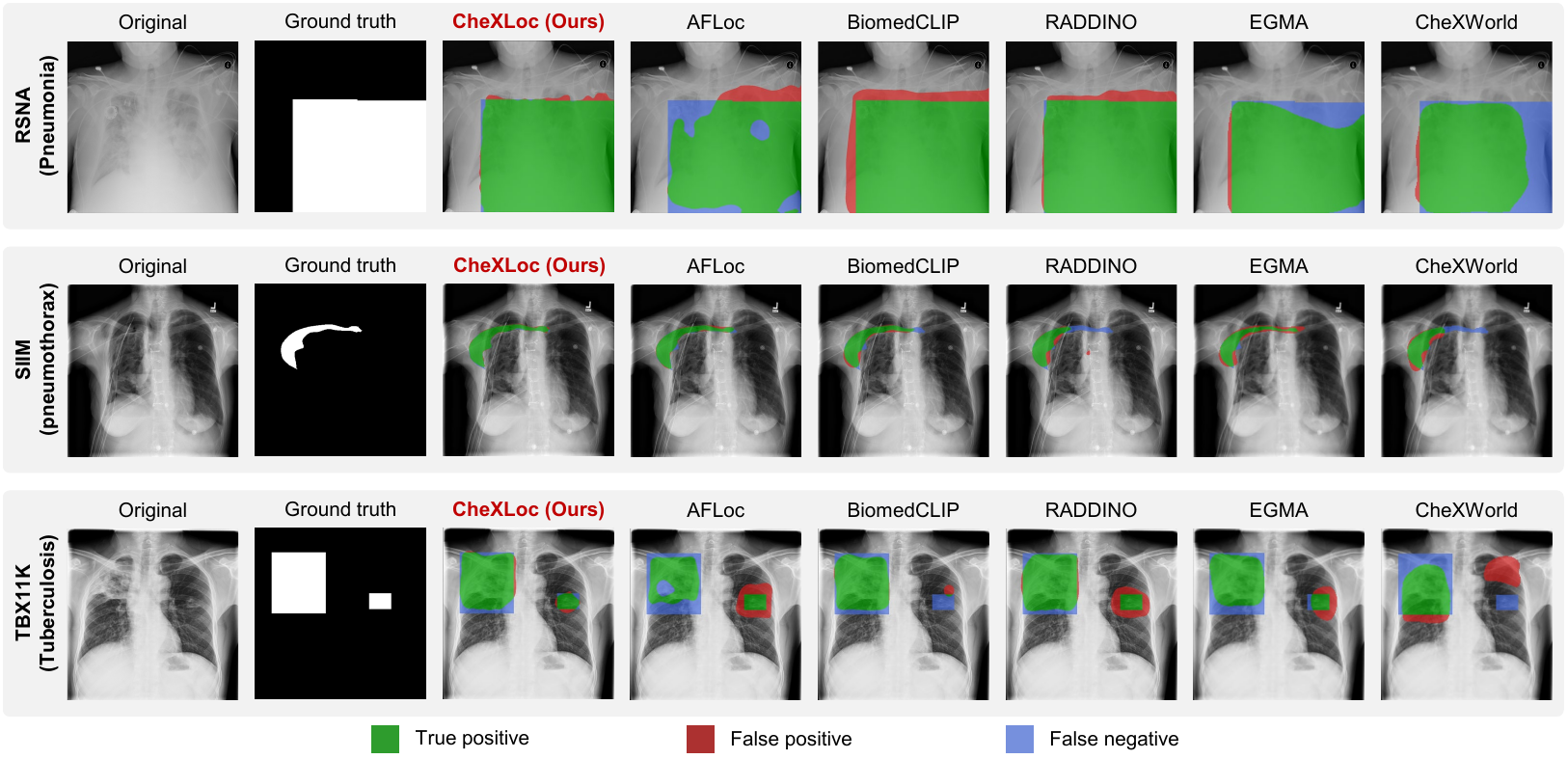}
	\caption{Representative lesion segmentation results for CheXLoc and existing medical vision--language models. Reference annotations are shown as box-derived masks for RSNA and TBX11K and pixel-level masks for SIIM. Pixel-level or box-derived lesion annotations were used exclusively for downstream supervised segmentation and were not used during CheXLoc pretraining.
	}
	\label{fig:segmentation}
\end{figure*}

% ===============zero-shot classification==========================
\begin{table}
	\centering
	\caption{Zero-shot AUROC (\%) on four independent validation cohorts (COVIDx \cite{2022-covidx}, SIIM \cite{siim-acr-pneumothorax-segmentation}, Shenzhen \cite{2023-shenzhencxr}, and TBX11K \cite{tbx11k-cvpr-2020}) for COVID-19, pneumothorax, and tuberculosis classification. }
	\setlength{\tabcolsep}{0.7mm}
	\label{tab:cls-zero-shot}
	\begin{tabular}{ccccc} 
		\toprule
		\textbf{ Model } & \textbf{ SIIM } & \textbf{ Shenzhen } & \textbf{ COVIDx } & \textbf{ TBX11K } \\ 
		\midrule
		AFLoc & 51.1{\scriptsize[50.3, 52.1]} & 76.2{\scriptsize[70.0, 82.1]} & 50.9{\scriptsize[46.1, 55.9]} & 70.2{\scriptsize[65.7, 74.8]} \\
		CheXWorld & 51.2{\scriptsize[49.6, 53.0]} & 43.7{\scriptsize[35.5, 51.9]} & 44.8{\scriptsize[40.0, 49.5]} & 48.3{\scriptsize[43.9, 53.0]} \\
		BiomedCLIP & 58.7{\scriptsize[55.1, 62.1]} & 89.0{\scriptsize[82.5, 94.2]} & 45.2{\scriptsize[39.8, 51.2]} & 78.7{\scriptsize[74.1, 83.0]} \\
		RADDINO & 50.0{\scriptsize[46.5, 53.7]} & 71.7{\scriptsize[63.1, 80.0]} & 50.3{\scriptsize[44.5, 55.6]} & 68.9{\scriptsize[63.8, 73.7]} \\
		EGMA & 50.5{\scriptsize[48.9, 52.4]} & 45.1{\scriptsize[39.2, 52.1]} & 50.0{\scriptsize[50.0, 50.0]} & 52.8{\scriptsize[48.5, 57.0]} \\
		\textbf{ CheXLoc } & \textcolor{red}{\textbf{60.0}}{\scriptsize[56.2, 63.9]} & \textcolor{red}{\textbf{97.4}}{\scriptsize[94.8, 99.4]} & \textcolor{red}{\textbf{52.8}}{\scriptsize[47.5, 58.8]} & \textcolor{red}{\textbf{91.3}}{\scriptsize[88.9, 93.4]} \\
		\bottomrule
	\end{tabular}
	
	\begin{minipage}{\linewidth}
		\vspace{1mm}
		\footnotesize
		\textit{Notes:} Values are reported as means with 95\% confidence intervals estimated from 1,000 bootstrap replicates.
	\end{minipage}
\end{table}

%
%Beyond report-generation quality, CheXLoc also has substantially fewer parameters and lower GPU memory requirements than the larger multimodal models evaluated (Fig.~\ref{fig:report-generation}b). Under the inference-time evaluation protocol described in Sec.~\ref{sec:inference}, CheXLoc required an average of 1.109~s per sample, compared with 4.894~s for MAIRA-2-7B \cite{bannur2024maira2groundedradiologyreport}. Together, these results suggest that CheXLoc achieves competitive report-generation performance while requiring substantially fewer parameters, lower GPU memory, and less inference time. Extended Data 
%%Table~\ref{tab:supp_inference_time} 
%Table~1 further presents the inference efficiency of CheXLoc across different batch sizes on the ReXRank-standardized MIMIC-CXR test subset.

\textbf{Generalization across External Datasets.}
We next evaluate CheXLoc on the external CheXpert Plus~\cite{chexpert-plus-2024} and ReXGradient~\cite{2025-ml4h-rexgradient} datasets without architectural changes (Fig.~\ref{fig:report-generation}c). The comparison includes general-purpose multimodal models (GPT-4V~\cite{op2023-gpt4v}), medical multimodal models (MedGemma-4B~\cite{medgemma}, MedVersa-7B~\cite{zhou2026medversa}, and RadFM-14B~\cite{wu2025towards-radfm}), and radiology-specific methods (Libra-3B~\cite{acl-2025-libra}, CheXagent-3B~\cite{chen2024-chexagent}, RadPhi3.5Vision-3.8B~\cite{radphi3.5vision}, and MAIRA-2-7B~\cite{bannur2024maira2groundedradiologyreport}). On CheXpert Plus, CheXLoc achieves the highest 1/RadCliQ-V1 score (0.870) while remaining competitive across the other evaluation metrics. On ReXGradient, a large-scale multi-institutional benchmark with 10,000 test studies, CheXLoc achieves the highest scores on all reported metrics: 1/RadCliQ-V1 (1.520), BLEU-2 (0.335), SembScore (0.537), RadGraph (0.358), and RATEScore (0.632). The consistent results across both external datasets suggest that the performance gains observed on MIMIC-CXR generalize beyond the source dataset, supporting the transferability of report-supervised visual representations across datasets and institutions.

%The preservation of performance across these datasets is notable because the models were evaluated outside the distribution used for report-supervised pretraining. In particular, the consistent improvements in clinically oriented metrics, such as RadGraph and RATEScore, suggest that the learned representations capture clinically relevant visual information rather than relying solely on dataset-specific lexical regularities. Together, these findings indicate that report-supervised pretraining can yield visual representations that retain performance across heterogeneous CXR data distributions rather than being restricted to source-domain reporting patterns.

\textbf{Zero-Shot Transfer across Institutions.}
To assess cross-institutional generalization, we evaluate CheXLoc on the IU X-ray dataset \cite{demner2016preparing-iu-xray} without fine-tuning ($n=2{,}955$; Tab.~\ref{tab:rg-iu-xray}). CheXLoc achieves the highest SembScore (0.565) and ranks second on the remaining four metrics, giving it the second-best mean rank after MedGemma-4B \cite{medgemma}. It uses approximately 7\% as many parameters as MedGemma-4B. These results show transfer to an independent institutional dataset without target-domain adaptation.

\begin{table}
	\centering
	\footnotesize
	\caption{Supervised classification performance (\%) in terms of F1-score on Shenzhen, SIIM, and TBX11K with 1\%, 10\%, and 100\% of the downstream training data.}
	\setlength{\tabcolsep}{0.10mm}
	\label{tab:cls-supervised}
	\begin{tabular}{cccccccccc} 
		\toprule
		\multirow{2}{*}{\textbf{ Model }} & \multicolumn{3}{c}{\textbf{ Shenzhen }} & \multicolumn{3}{c}{\textbf{ SIIM }} & \multicolumn{3}{c}{\textbf{ TBX11K }} \\ 
		\cmidrule(lr){2-4}\cmidrule(lr){5-7}\cmidrule(lr){8-10}
		& \textbf{ 1\% } & \textbf{ 10\% } & \textbf{ 100\% } & \textbf{ 1\% } & \textbf{ 10\% } & \textbf{ 100\% } & \textbf{ 1\% } & \textbf{ 10\% } & \textbf{ 100\% } \\ 
		\midrule
		AFLoc & 59.8 & 85.5 & 93.4 & 69.4 & 86.1 & \textcolor{red}{\textbf{ 98.7 }} & 22.5 & 87.5 & 96.6 \\
		CheXWorld & 79.4 & 83.6 & 84.3 & 20.0 & 48.5 & 83.9 & 49.2 & 50.0 & 70.7 \\
		BiomedCLIP & 28.9 & 79.5 & 92.8 & 72.6 & 80.6 & 98.0 & 38.3 & 69.8 & 94.9 \\
		RADDINO & 65.2 & 88.3 & 95.7 & 51.2 & 86.6 & 97.1 & 16.8 & 88.3 & 96.6 \\
		EGMA & 87.1 & 89.1 & 89.4 & 60.8 & 81.8 & 98.1 & 17.0 & 82.6 & 94.3 \\
		\textbf{CheXLoc} & \textcolor{red}{\textbf{ 91.2 }} & \textcolor{red}{\textbf{ 94.3 }} & \textcolor{red}{\textbf{ 97.1 }} & \textcolor{red}{\textbf{ 86.8 }} & \textcolor{red}{\textbf{ 87.5 }} & 98.3 & \textcolor{red}{\textbf{ 64.9 }} & \textcolor{red}{\textbf{ 88.5 }} & \textcolor{red}{\textbf{ 97.9 }} \\
		\bottomrule
	\end{tabular}
\end{table}

%==================image dependence=====================
\begin{figure*}
	\centering
	\includegraphics[width=1\linewidth]{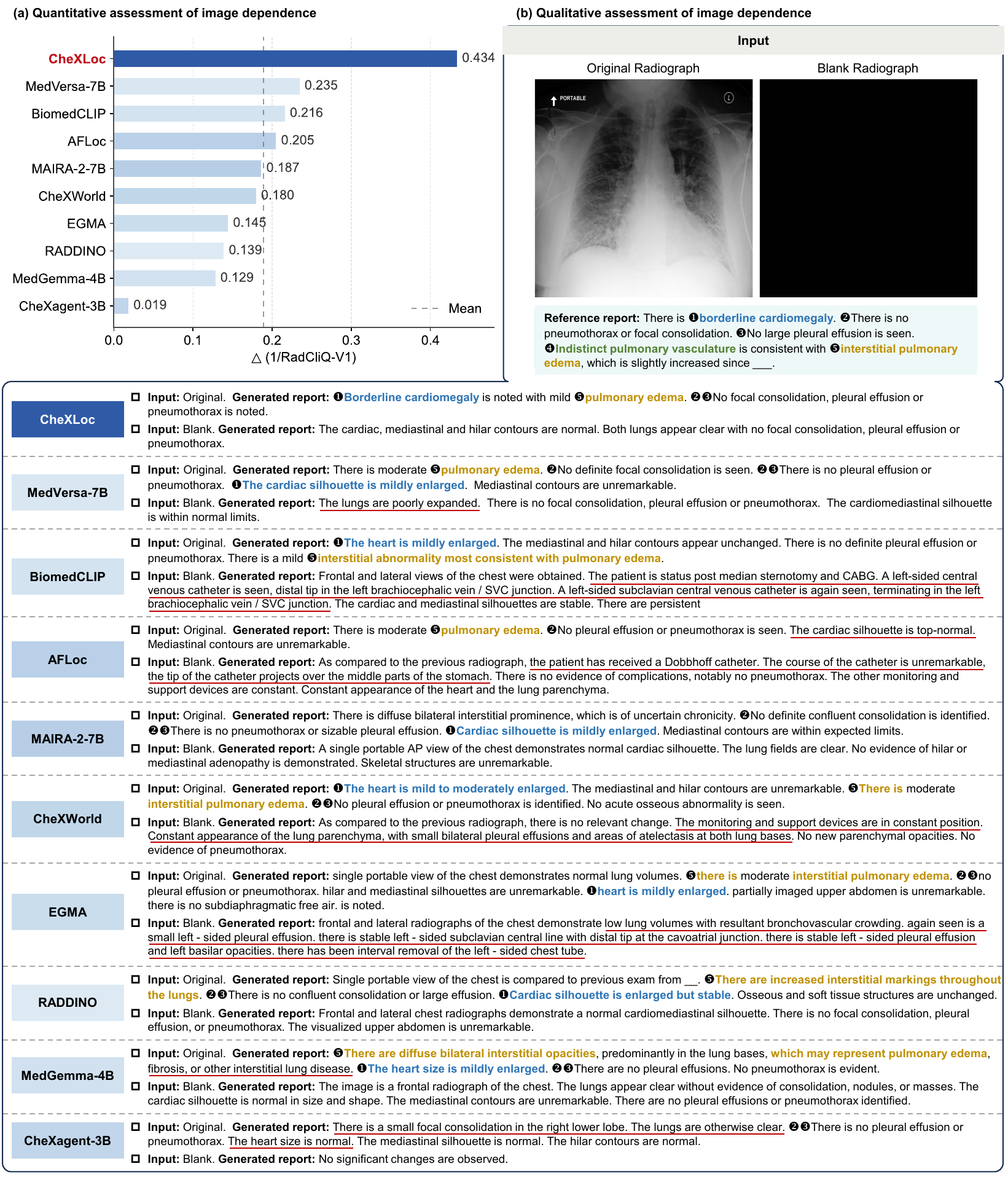}
	\caption{
		Assessing image dependence in report generation.
		\textbf{a,} Quantitative assessment of visual dependence based on 
		$\bigtriangleup (1/\text{RadCliQ-V1} )$ between reports generated from original and blank radiographs on the ReXRank-MIMIC-CXR ($n=2{,}347$). Larger positive values indicate a greater reliance on visual information.
		\textbf{b,} Qualitative comparison of the generated reports. Numbered sentences indicate correct matches to the reference report, while colored text highlights correctly generated abnormal findings or devices. \textcolor{red}{\underline{Clinically significant errors}} are highlighted in \textcolor{red}{\underline{red underlining}}.
	}
	\label{fig:dependance}
\end{figure*}
%CheXLoc generates reports that closely align with the reference reports when conditioned on original radiographs, while producing fewer hallucinated abnormalities and shifting towards normal descriptions under blank inputs

\subsection{Transferability across Diagnostic Tasks}

\textbf{Zero-shot Disease Classification without Disease-Specific Supervision.}
As illustrated in Tab.~\ref{tab:cls-zero-shot}, CheXLoc achieves the highest AUROC on all four benchmarks without disease-specific training or adaptation. The largest margins occur on Shenzhen and TBX11K, where AUROCs reach 97.4\% (95\% CI: 94.8\%--99.4\%) and 91.3\% (95\% CI: 88.9\%--93.4\%), respectively, surpassing the strongest baselines by 8.4 and 12.6 percentage points. CheXLoc also obtains the highest AUROCs on SIIM (60.0\%) and COVIDx (52.8\%). These results indicate transfer to zero-shot disease classification across the evaluated datasets.

\textbf{Supervised Transfer under Limited Labeled Data.}
As shown in Tab.~\ref{tab:cls-supervised}, CheXLoc performs particularly well when downstream labels are limited. With 1\% of the training data, it achieves the highest F1-scores on all three datasets: 91.2\% on Shenzhen, 86.8\% on SIIM, and 64.9\% on TBX11K. The advantage persists with 10\% of the data, yielding scores of 94.3\%, 87.5\%, and 88.5\%, respectively.  With the full training set, CheXLoc achieves the highest scores on Shenzhen (97.1\%) and TBX11K (97.9\%) and remains competitive on SIIM (98.3\%), slightly below AFLoc (98.7\%). These results indicate that the learned representations are transferable, particularly in low-label settings.

\subsection{Lesion Segmentation across Thoracic Diseases}
We evaluate transfer to lesion segmentation on RSNA, SIIM, and TBX11K. As reported in Tab.~\ref{tab:seg-supervised}, CheXLoc achieves mean Dice scores of 80.4\%, 73.5\%, and 96.5\%, respectively, ranking first on all three benchmarks. It exceeds the second-best methods by 1.2 percentage points on RSNA and 10.4 points on SIIM, while remaining competitive on TBX11K. The results indicate that report-supervised pretraining transfers to segmentation across the evaluated diseases and annotation formats.

Qualitative examples in Fig.~\ref{fig:segmentation} are consistent with the quantitative results. On RSNA, CheXLoc produces more spatially complete predictions for pneumonia lesions; on TBX11K, it identifies subtle tuberculosis lesions missed in part by competing methods; and on SIIM, its masks more closely follow irregular pneumothorax boundaries. These examples suggest that the learned representations transfer across the evaluated diseases and annotation formats, from box-derived to pixel-level masks.

% =========================segmentation performance ===============================
\begin{table}
	\centering
	\caption{lesion segmentation performance (\%) in terms of mean Dice score on the RSNA, SIIM, and TBX11K benchmarks.}
	\setlength{\tabcolsep}{0.8mm}
	\label{tab:seg-supervised}
	\begin{tabular}{cccc} 
		\toprule
		\textbf{ Model } & \textbf{ RSNA } & \textbf{ SIIM } & \textbf{ TBX11K } \\ 
		\midrule
		AFLoc \cite{2026-nbe-afloc} & 69.7 [68.3, 70.9] & 63.1 [61.0, 65.0] & 95.1 [94.1, 96.1] \\
		CheXWorld \cite{2025-cvpr-chexworld} & 78.2 [77.0, 79.4] & 25.2 [22.9, 27.4] & 92.6 [91.3, 93.8] \\
		BiomedCLIP \cite{zhang2025biomedclipmultimodalbiomedicalfoundation} & 77.3 [76.1, 78.4] & 62.5 [60.7, 64.4] & 95.3 [94.3, 96.2] \\
		RADDINO \cite{2024-rad-dino-nmi} & 79.2 [78.1, 80.3] & 59.2 [57.4, 61.0] & 96.2 [95.3, 96.9] \\
		EGMA \cite{ma2024-nips-eye-egma} & 78.1 [76.9, 79.3] & 55.8 [53.8, 57.7] & 95.1 [94.2, 96.1] \\
		\textbf{ CheXLoc (Ours)} & \textcolor{red}{\textbf{ 80.4}} [79.3, 81.5] & \textcolor{red}{\textbf{ 73.5}} [71.7, 75.3] & \textcolor{red}{\textbf{ 96.5}} [95.7, 97.2] \\
		\bottomrule
	\end{tabular}
	
	\begin{minipage}{\linewidth}
		\vspace{1mm}
		\footnotesize
		\textit{Notes:} Values are reported as means with 95\% confidence intervals estimated from 1,000 bootstrap replicates.
	\end{minipage}
\end{table}

\begin{figure*}
	\centering
	\includegraphics[width=1\linewidth]{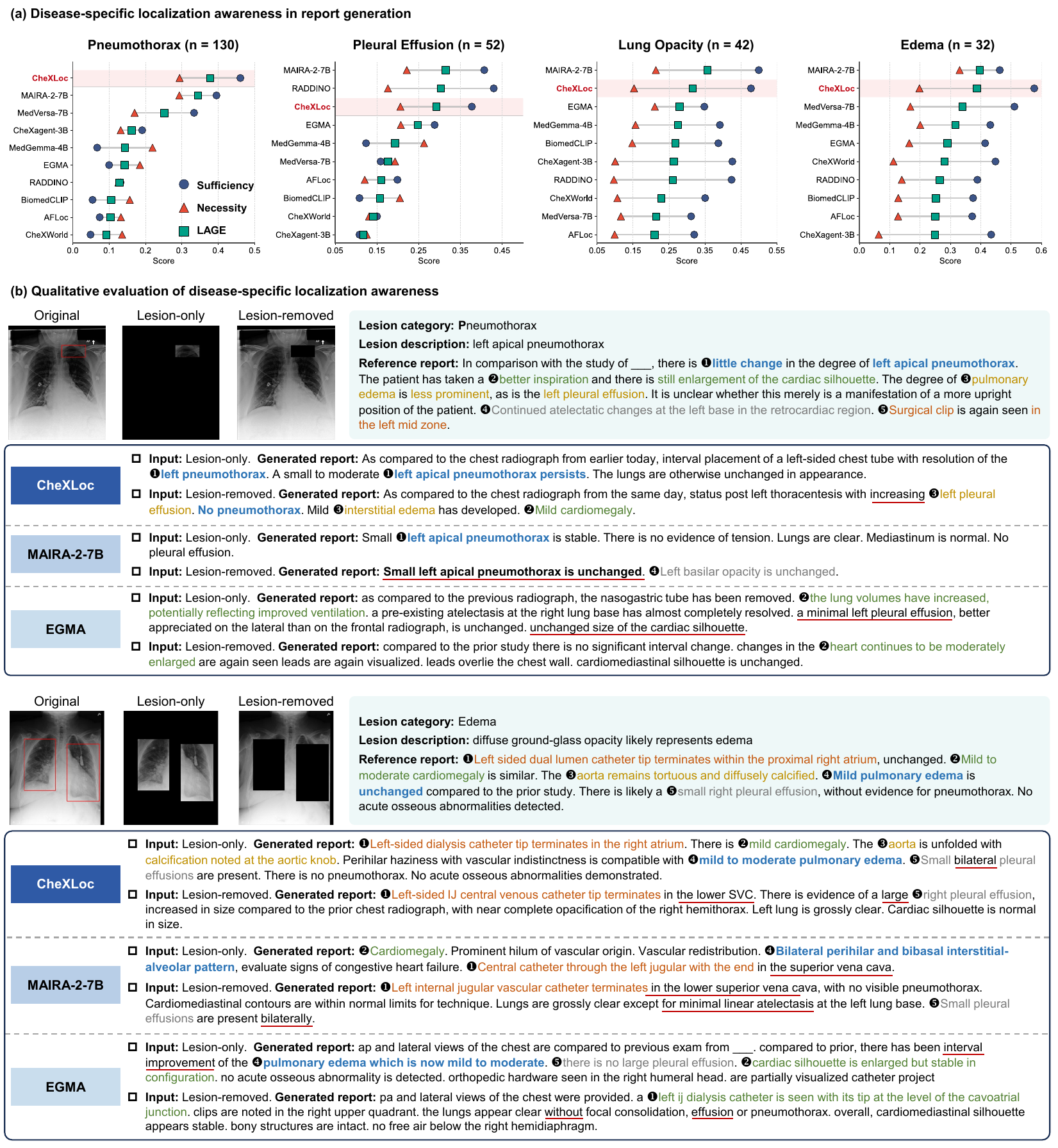}
	\caption{
		Disease-specific localization awareness in chest X-ray report generation.
		\textbf{a,} Quantitative comparison of nine medical foundation models across four thoracic diseases in terms of Sufficiency, Necessity, and the integrated LAGE score.
		\textbf{b,} Qualitative comparison of generated reports under lesion-only and lesion-removed interventions. Numbered sentences indicate correct matches to the reference report, while colored text highlights correctly generated abnormal findings or devices. \textcolor{red}{\underline{Clinically significant errors}} are indicated by \textcolor{red}{\underline{red underlining}}.
	}
	\label{fig:localization}
\end{figure*}

%\subsection{Sensitivity of Report Generation to Visual and Disease-Specific Evidence}
%We evaluated disease-specific visual evidence at two complementary levels. We first quantified the dependence of report generation on visual input by removing the radiograph and then examined whether generated findings responded to disease-specific pathological regions using lesion-level interventions.

\subsection{Sensitivity of Report Generation to Visual Input}

\textbf{Quantitative Analysis.}
To quantify the contribution of visual input to report generation, we evaluate CheXLoc on the ReXRank-MIMIC ($n=2{,}347$) using either the original radiograph or a blank image, with all other generation settings fixed. We define visual dependence as the difference in report quality between these conditions:
\begin{align}
	\bigtriangleup R =  R_{\text{Original} } - R_{\text{Blank} },
\end{align}
where $R$ denotes 1/RadCliQ-V1. Larger positive values indicate a greater reduction in report quality upon removal of visual input. As shown in Fig.~\ref{fig:dependance}a, CheXLoc exhibits the largest difference ($\bigtriangleup \text{1/RadCliQ-V1}=0.434$), compared with 0.019--0.235 across the other models. This result suggests greater sensitivity of its generated reports to the presence of visual input.

\textbf{Qualitative Analysis.}
We compare reports generated from the original and blank radiographs to examine how visual input affects report content (Fig.~\ref{fig:dependance}b). Given the original radiograph, CheXLoc correctly captures the borderline cardiomegaly and edema described in the reference report without overstating their severity. Although several baseline models also identify these abnormalities, they tend to overestimate the severity of cardiomegaly, for instance, by describing borderline cardiomegaly as mild cardiomegaly. RADDINO \cite{2024-rad-dino-nmi} additionally generates \textit{stable cardiomegaly}, which is not supported by the reference report, whereas AFLoc \cite{2026-nbe-afloc}, MAIRA-2-7B \cite{bannur2024maira2groundedradiologyreport}, and CheXagent-3B \cite{chen2024-chexagent} omit one or more clinically relevant findings.

When the radiograph is replaced with a blank image, CheXLoc, MAIRA-2-7B \cite{bannur2024maira2groundedradiologyreport}, RADDINO \cite{2024-rad-dino-nmi}, MedGemma-4B \cite{medgemma}, and CheXagent-3B \cite{chen2024-chexagent} generate reports without asserting abnormal findings. Ideally, a model would identify the blank image as invalid and request a valid radiograph; however, this behavior is not explicitly supervised during training. Because a blank radiograph provides no visual evidence of abnormal findings, generating a normal report is a reasonable response. In contrast, MedVersa-7B, BiomedCLIP, AFLoc, CheXWorld, and EGMA \cite{ma2024-nips-eye-egma} generate abnormal findings despite the absence of corresponding visual evidence. These observations are consistent with the quantitative results and suggest that CheXLoc is more sensitive to visual input.

\begin{figure*}
	\centering
	\includegraphics[width=1\linewidth]{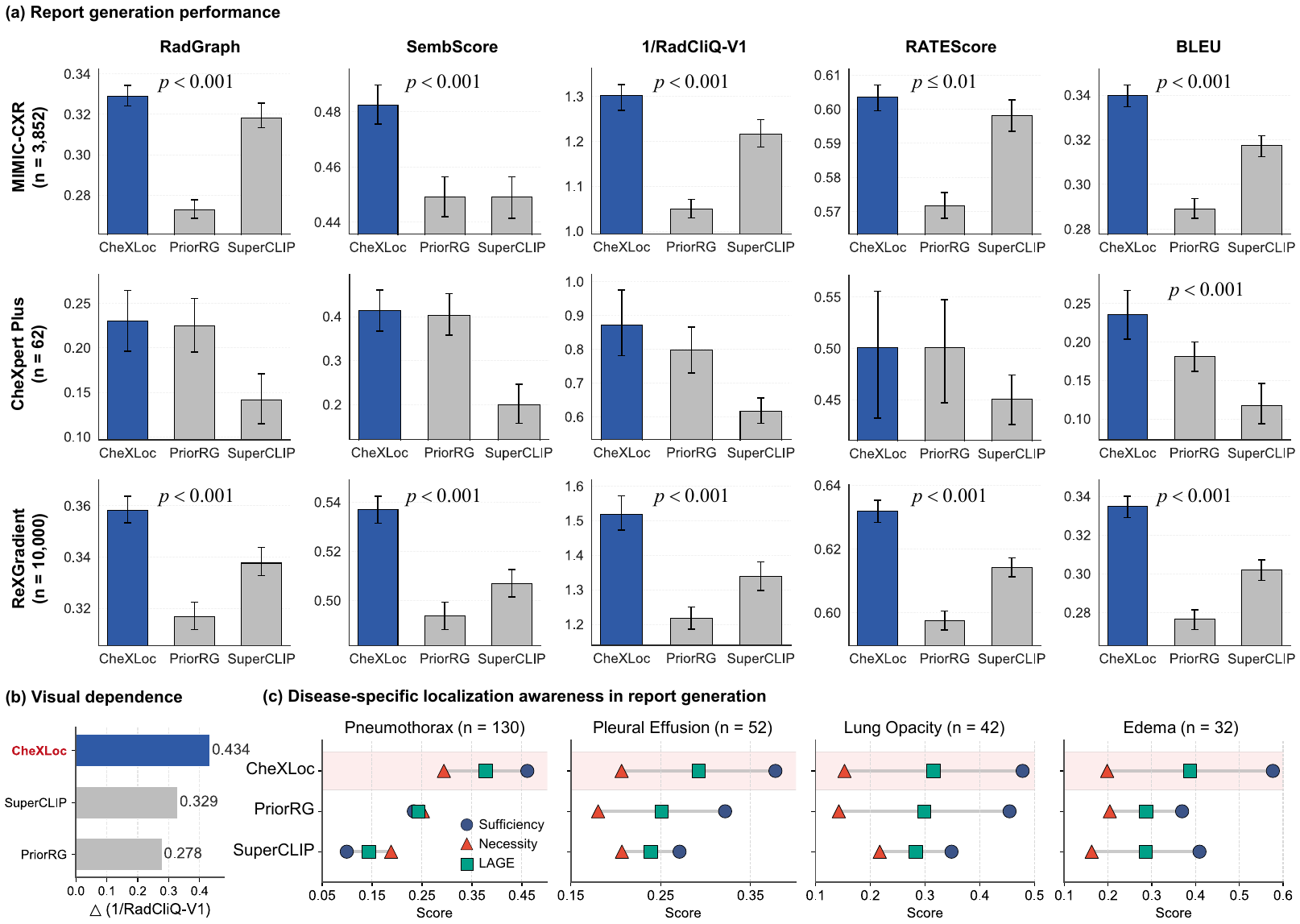}
	\caption{
		Comparison of global and token-weighted visual--textual alignment strategies. We compare CheXLoc with PriorRG \cite{PriorRG}, a global alignment baseline, and SuperCLIP \cite{zhao2025-superclip}, a frequency-based token-weighting baseline, across report generation, sensitivity to visual input, and disease-specific sensitivity to localized evidence.
		\textbf{a,} Report generation performance on internal and external datasets.
		\textbf{b,} Sensitivity to visual input, measured by $\bigtriangleup (1/\text{RadCliQ-V1})$ between reports generated from original and blank radiographs ($n=2{,}347$).
		\textbf{c,} Disease-specific sensitivity to localized evidence across four thoracic diseases, evaluated using Sufficiency, Necessity, and LAGE.
	}
	\label{fig:ablation-study-superclip}
\end{figure*}

\subsection{Sensitivity of Report Generation to Disease-Specific Evidence}
\textbf{Quantitative Analysis.}
We evaluate nine medical foundation models on MS-CXR~\cite{PhysioNet-ms-cxr-1.1.0} using LAGE (Sec.~\ref{metrics}) to assess the sensitivity of generated findings to lesion-level perturbations. LAGE combines lesion-only and lesion-removed interventions.

As shown in Fig.~\ref{fig:localization}a, CheXLoc achieves the highest LAGE score for pneumothorax (0.377), with sufficiency and necessity scores of 0.461 and 0.294, respectively. Most baselines obtain LAGE scores below 0.200. For pleural effusion, lung opacity, and edema, CheXLoc ranks second or third, with LAGE scores within 0.023, 0.041, and 0.010 of the respective best-performing model, MAIRA-2-7B \cite{bannur2024maira2groundedradiologyreport}. Notably, MAIRA-2-7B, MedVersa-7B \cite{zhou2026medversa}, and EGMA \cite{ma2024-nips-eye-egma} use spatial supervision during pretraining, whereas CheXLoc relies solely on image--report pairs. Despite the absence of explicit spatial supervision, CheXLoc achieves competitive LAGE performance across all four findings, suggesting that report supervision provides a useful weak signal for learning visual representations sensitive to disease-specific regions.

\textbf{Qualitative Analysis.}
We compare the responses of representative models to lesion-only and lesion-removed inputs for pneumothorax and edema (Fig.~\ref{fig:localization}b). For pneumothorax (Case 1), CheXLoc and MAIRA-2-7B \cite{bannur2024maira2groundedradiologyreport} correctly generate \textit{left apical pneumothorax} from the lesion-only image, whereas EGMA \cite{ma2024-nips-eye-egma} fails to recover this finding. When the lesion is removed, CheXLoc no longer reports pneumothorax while preserving the other supported abnormalities. In contrast, MAIRA-2-7B continues to report pneumothorax despite the removal of its corresponding visual evidence, whereas EGMA omits \textit{left pleural effusion} from the lesion-only input.

For edema (Case 2), all three models recover the edema-related finding from the lesion-only input. Because the annotated edema region covers much of the thorax and upper abdomen, other abnormalities remain visible. CheXLoc also identifies additional findings, including \textit{catheter tip terminating in the right atrium}, \textit{aortic unfolding with calcification at the aortic knob}, and \textit{small pleural effusion}, whereas MAIRA-2-7B and EGMA fail to capture these findings. After lesion removal, all three models suppress the edema-related finding. These examples illustrate that CheXLoc responds to the presence and removal of targeted lesion evidence while retaining other findings supported by the remaining visual information.

Overall, these analyses indicate that CheXLoc's generated findings respond to disease-specific image perturbations, particularly for focal abnormalities such as pneumothorax, while retaining sensitivity to other findings supported by broader regions.

\subsection{Ablation Studies}
We conduct controlled ablation experiments to evaluate the contributions of the two alignment objectives and investigate key design choices in entity-guided local patch--token alignment. Specifically, we examine: (i) whether entity-guided local patch--token alignment further improves performance over global image--report alignment; (ii) whether incorporating clinical entities explicitly into token importance estimation provides advantages over frequency-based token weighting; (iii) the individual and complementary contributions of global image--report alignment and entity-guided local patch--token alignment; and (iv) the effects of different design choices in the entity-guided local patch--token alignment objective.

% chexloc vs. priorrg
\textbf{Effect of Entity-Guided Local Patch--Token Alignment.}
We compare CheXLoc with our PriorRG \cite{PriorRG} along three complementary dimensions: in-domain and external report generation, visual dependence, and sensitivity to disease-specific regions (Fig.~\ref{fig:ablation-study-superclip}). Both methods share the same encoders, report generator, pretraining data, optimization schedule, and training budget; they differ only in their alignment objectives. PriorRG relies solely on global image--report alignment, whereas CheXLoc additionally employs entity-guided local patch--token alignment to associate clinically informative tokens with localized visual representations.

For report generation, CheXLoc consistently outperforms PriorRG on the full MIMIC-CXR test set ($n=3{,}852$) and the external ReXGradient dataset ($n=10{,}000$), achieving statistically significant differences across all evaluated lexical and clinically oriented metrics ($p<0.001$). Notably, 1/RadCliQ-V1 increases by 25.2 and 30.1 percentage points on MIMIC-CXR and ReXGradient, respectively. On the smaller external CheXpert Plus cohort ($n=62$), CheXLoc achieves the highest overall performance, although the differences over PriorRG do not reach statistical significance.

These improvements in report generation are accompanied by greater sensitivity to visual evidence. CheXLoc exhibits substantially higher visual dependence than PriorRG (0.434 versus 0.278) and achieves higher LAGE scores across all four evaluated diseases: edema (0.387 versus 0.287), lung opacity (0.316 versus 0.299), pleural effusion (0.292 versus 0.251), and pneumothorax (0.377 versus 0.243). These results suggest that adding entity-guided local patch--token alignment to global image--report alignment improves report generation and sensitivity to disease-specific visual evidence.

%\begin{figure}[!h]
%	\centering
%	\includegraphics[width=1\linewidth]{figures/figure-ablation-study-reweight.pdf}
%	\caption{
%		\textbf{Ablation analysis of global image--report alignment ($\mathcal{L}_{\text{global}}$) and entity-guided local patch--token alignment ($\mathcal{L}_{\text{local}}$) on the full MIMIC-CXR test set ($n=3{,}852$).}  
%		\textbf{a,} Comparison of $\mathcal{L}_{\text{global}}$, $\mathcal{L}_{\text{local}}$, and CheXLoc ($\mathcal{L}_{\text{global}}+\mathcal{L}_{\text{local}}$);
%		\textbf{b,} Disentangling entity-guided local patch--token alignment. GPV denotes globally pooled visual features used for probability prediction. 	
%	}
%	\label{fig:ablation-study-reweight}
%\end{figure}

% =============================ablation study for each components=============
\begin{table*}
	\centering
	\caption{
			Ablation study of global image--report alignment ($\mathcal{L}_{\text{global}}$) and entity-guided local patch--token alignment ($\mathcal{L}_{\text{local}}$) on the full MIMIC-CXR test set ($n=3{,}852$), including comparisons of individual and joint alignment objectives and an analysis of the design choices in entity-guided local patch--token alignment.
		}
	\setlength{\tabcolsep}{0.9mm}
	\label{tab:ablation-study-components}
	\begin{tabular}{cccccccccc} 
		\toprule
		\textbf{ Settings } & \textbf{ 1/RadCliQ-V1$\uparrow$ } & \textbf{ SembScore$\uparrow$} & \textbf{ RadGraph$\uparrow$ } & \textbf{ BLEU-2$\uparrow$ } & \textbf{ BLEU-4$\uparrow$ } & \textbf{ $^5$Mi-F1$\uparrow$ } & \textbf{ $^5$Ma-F1$\uparrow$ } & \textbf{ $^{14}$Mi-F1$\uparrow$ } & \textbf{ $^{14}$Ma-F1$\uparrow$ } \\ 
		\midrule
		$\mathcal{L}_{\text{global}}$ & 1.050 & 0.449 & 0.272 & 0.290 & 0.175 & 0.562 & 0.496 & 0.511 & 0.376 \\
		$\mathcal{L}_{\text{local}}$ & 0.960 & 0.423 & 0.252 & 0.262 & 0.151 & 0.524 & 0.444 & 0.482 & 0.337 \\
		Glob. Pooled Vis. Feat. & 1.265 & 0.460 & 0.328 & 0.306 & 0.212 & 0.481 & 0.408 & 0.447 & 0.297 \\
		No Reweighting & 1.242 & 0.446 & 0.326 & 0.306 & 0.213 & 0.462 & 0.392 & 0.425 & 0.286 \\
		Fixed Weights & 1.247 & 0.450 & 0.326 & 0.303 & 0.211 & 0.484 & 0.409 & 0.435 & 0.288 \\
		Uniform Weights & 1.238 & 0.448 & 0.327 & 0.306 & 0.213 & 0.482 & 0.410 & 0.432 & 0.294 \\
		\textbf{CheXLoc (Ours)} & \textcolor{red}{\textbf{1.301}} & \textcolor{red}{\textbf{0.483}} & \textcolor{red}{\textbf{0.329}} & \textcolor{red}{\textbf{0.340}} & \textcolor{red}{\textbf{0.226}} & \textcolor{red}{\textbf{0.586}} & \textcolor{red}{\textbf{0.509}} & \textcolor{red}{\textbf{0.534}} & \textcolor{red}{\textbf{0.385}} \\
		\bottomrule
	\end{tabular}
\end{table*}

% chexloc vs. superclip
\textbf{Effect of Entity-Informed Token Weighting.}
We compare CheXLoc with SuperCLIP \cite{zhao2025-superclip} (Fig.~\ref{fig:ablation-study-superclip}) to isolate the effect of clinical entity guidance in token importance estimation. SuperCLIP estimates token importance using inverse document frequency, whereas CheXLoc derives token weights from learnable importance weights assigned to clinical entity types. All other model components and training configurations are kept identical to isolate the effect of the weighting strategy.

On the full MIMIC-CXR test set ($n=3{,}852$), CheXLoc significantly outperforms SuperCLIP across the evaluated lexical and clinically oriented metrics (all $p\leq0.01$). Specifically, RadGraph increases from 0.318 to 0.329, 1/RadCliQ-V1 from 1.216 to 1.303, and SembScore from 0.449 to 0.482. Similar improvements are observed on the external CheXpert Plus and ReXGradient datasets, where CheXLoc achieves higher scores across all evaluated metrics (all $p<0.001$).

The benefits of clinical entity guidance also extend to visual evidence utilization. CheXLoc exhibits greater visual dependence than SuperCLIP (0.434 versus 0.329) and achieves higher LAGE scores for all four evaluated diseases. These findings suggest that, in the evaluated settings, clinical entity guidance provides benefits over frequency-based token weighting for report generation and sensitivity to disease-specific visual evidence.

\textbf{Contributions of Global Image--Report Alignment and Entity-Guided Local Patch--Token Alignment.}
To separate the contributions of the two alignment objectives, we evaluate variants trained with either $\mathcal{L}_{\text{global}}$ or $\mathcal{L}_{\text{local}}$ alone on the full MIMIC-CXR test set ($n=3{,}852$; Tab.~\ref{tab:ablation-study-components}). The global-only variant consistently outperforms the local-only variant across all nine metrics, indicating the importance of holistic image--report correspondence. Nevertheless, neither objective alone achieves the performance of their joint optimization. CheXLoc yields the best results across all evaluated metrics, including 1/RadCliQ-V1 (1.301), SembScore (0.483), RadGraph (0.329), BLEU-2 (0.340), and BLEU-4 (0.226), as well as the highest micro- and macro-F1 scores in both the 5- and 14-observation settings. These findings demonstrate that global and local alignment provide complementary supervision, with the former capturing holistic cross-modal correspondence and the latter establishing finer-grained associations between visual regions and clinically informative tokens.

\textbf{Disentangling Entity-Guided Local Patch--Token Alignment.}
We next examine the key design choices of entity-guided local patch--token alignment (Tab.~\ref{tab:ablation-study-components}). Replacing patch-level features with globally pooled visual features (\textit{Glob. Pooled Vis. Feat.}) consistently degrades performance, with particularly pronounced reductions in $^5$Mi-F1 (0.586 to 0.481) and $^{14}$Mi-F1 (0.534 to 0.447). This suggests that preserving spatial information is important for establishing fine-grained visual--textual associations.

We further investigate the contribution of token importance reweighting by removing the reweighting operation while retaining the binary cross-entropy loss in Eq.~(\ref{eq:reweight}), yielding the \textit{No Reweighting} variant. Without reweighting, the variant identifies clinical tokens but assigns no differential importance to them. Performance decreases on all nine metrics, including SembScore (0.483 to 0.446), BLEU-2 (0.340 to 0.306), and $^5$Mi-F1 (0.586 to 0.462). Thus, identifying clinical tokens alone is insufficient; weighting their relative importance provides additional benefit.

Finally, we examine the initialization and adaptation of the importance mapping $\boldsymbol{\alpha}(\cdot)$. \textit{Uniform Weights} initializes all weights in the importance mapping to 1 and allows them to be optimized during training. In contrast, \textit{Fixed Weights} initializes the weights using expert-informed values and keeps them frozen throughout training. As shown in Tab.~\ref{tab:ablation-study-components}, both \textit{Uniform Weights} and \textit{Fixed Weights} underperform CheXLoc across all evaluated metrics. \textit{Uniform Weights} yields lower SembScore (0.448), BLEU-4 (0.213), and $^{14}$Mi-F1 (0.432), while \textit{Fixed Weights} results in corresponding scores of 0.450, 0.211, and 0.435. The results suggest that combining expert-informed initialization with data-driven adaptation is beneficial for entity-importance estimation.

%We further examine the effect of expert-informed initialization by initializing the importance mapping $\boldsymbol{\alpha}(\cdot)$ uniformly across entity types and allowing it to adapt during training, yielding the \textit{Uniform Weights} variant. Compared with CheXLoc, this variant exhibits lower performance across all nine evaluated metrics. In particular, 1/RadCliQ-V1 decreases from 1.301 to 1.238, BLEU-4 from 0.226 to 0.213, and $^{14}$Mi-F1 from 0.534 to 0.432. These results suggest that expert-informed initialization provides a more effective prior for learning entity importance than uniform initialization.
%
%
%Finally, we assess the benefit of adapting the expert-informed importance mapping $\boldsymbol{\alpha}(\cdot)$ during training by keeping its initialized values fixed, yielding the \textit{Fixed Weights} variant. Although this variant retains the expert-informed prior, it consistently underperforms CheXLoc across all nine metrics. Specifically, 1/RadCliQ-V1 decreases from 1.301 to 1.247, BLEU-2 from 0.340 to 0.303, and $^{14}$Mi-F1 from 0.534 to 0.435. These results indicate that expert-informed initialization alone is insufficient and that allowing the importance mapping to adapt during training further improves the estimation of entity importance.

\section{Discussion}
Our results show that clinical entities extracted from radiology reports can provide effective weak supervision for learning locally aligned visual representations without explicit spatial annotations. The ablation results demonstrate the complementary roles of global image--report alignment and entity-guided local patch--token alignment. The performance differences among the local alignment variants further indicate the contributions of patch-level features and token reweighting. In particular, the results support the use of learnable entity-type weights, rather than uniform or fixed weights, for deriving token-level importance. They also suggest that expert-informed initialization and data-driven adaptation provide complementary benefits.

CheXLoc transfers across report generation, disease classification, and lesion segmentation, with additional benefits in disease classification under limited supervision. This suggests that report-derived supervision can support reusable visual representations when task-specific annotations are scarce. Perturbation analyses further show that the learned representations are sensitive to disease-related visual evidence, with stronger effects for focal abnormalities such as pneumothorax than for findings with broader spatial distributions, such as edema and pleural effusion. However, these results reflect sensitivity to controlled image modifications and do not establish explicit spatial correspondence between individual clinical entities and image regions.

This study has several limitations. First, automated entity extraction may introduce errors or noise into token importance estimation. Second, our evaluation is limited primarily to chest radiographs and English-language reports; generalization to other modalities, anatomical regions, and languages remains to be studied. Third, perturbation-based evaluation provides indirect evidence of localization awareness. LAGE assumes that removing an annotated disease region eliminates the corresponding finding while leaving other findings unchanged, which may not hold when the annotated region is sufficiently large to encompass other non-target abnormalities. Future work will examine entity-level visual correspondence more directly and extend the framework to additional imaging modalities and clinical settings.

\section{Conclusion}

In this work, we introduced CheXLoc, a report-supervised vision--language framework for learning locally aligned visual representations from chest radiographs without explicit spatial annotations. CheXLoc derives entity-informed token weights from learnable entity-type importance and incorporates them into local patch--token alignment, complementing global image--report alignment. Extensive experiments demonstrate its effectiveness across report generation, disease classification, and lesion segmentation, with particularly strong performance in classification under limited downstream supervision. Furthermore, image-level and lesion-level perturbation analyses provide complementary evidence of the model's sensitivity to clinically relevant visual information. These findings highlight clinical entity-guided report supervision as a scalable source of weak supervision for learning locally aligned visual representations without explicit spatial annotations during pretraining.

\section{Acknowledgments}
This work was jointly supported by the National Natural Science Foundations of China (Nos. 62272364, 62672390, and 62677054), the Provincial Key Research and Development Program of Shaanxi (No. 2025CG-GJHX-19), and the Higher Education Science Research Planning Project of China Association of Higher Education (No. 24PG0101).

\bibliographystyle{IEEEtran}
\bibliography{main-details}

%Biography text here without a photo.

\end{document}